\documentclass[journal]{IEEEtran}
\usepackage{amsmath}
\usepackage{amsfonts}
\usepackage{bbm}
\usepackage{algorithmic}
\usepackage{float}
\usepackage[linesnumbered,ruled]{algorithm2e}
\usepackage{array}
\usepackage{textcomp}
\usepackage{stfloats}
\usepackage{url}
\usepackage{verbatim}
\usepackage{graphicx}
\usepackage{subcaption}
\usepackage{cite}
\usepackage{amssymb}
\usepackage{wrapfig}
\usepackage{physics}
\usepackage{xcolor}
\usepackage{xspace}
\usepackage{ragged2e}
\usepackage{booktabs,makecell,multirow,tabularx}
\usepackage[colorlinks,linkcolor=blue,anchorcolor=blue,citecolor=blue]{hyperref}
\usepackage{cleveref}

\ifCLASSINFOpdf
\else
   \usepackage[dvips]{graphicx}
\fi
\usepackage{url}

\usepackage{graphicx}

\usepackage{booktabs}
\usepackage{multirow}
\usepackage{rotating}
\usepackage{mathrsfs,amssymb,amsmath, amsfonts,amsthm}

\begin{document}

\title{Robust Duality Learning for Unsupervised Visible-Infrared Person Re-Identiﬁcation}
\author{Yongxiang Li, Yuan Sun, Yang Qin, Dezhong Peng, Xi Peng and Peng Hu
\thanks{Manuscript received 27 July 2024; revised 26 December 2024;
accepted 16 January 2025. Date of publication xx xx 2025; date
of current version xx xx 2025. This work was supported in part by the National Key R\&D Program of China under Grant 2024YFB4710604; in part by the National Natural Science Foundation of China (NSFC) under Grants 62472295, 62176171, U21B2040, and 62372315; in part by the Sichuan Science and Technology Planning Project under Grants 2024NSFTD0047 and 2024NSFTD0038; in part by the System of Systems and Artificial Intelligence Laboratory Pioneer Fund Grant; in part by the Fundamental Research Funds for the Central Universities under Grants CJ202303 and CJ202403; and in part by the Sichuan Science and Technology Planning Projects under Grants 2024NSFTD0049, 2024ZDZX0004, 2024YFHZ0144, and 2024YFHZ0089. \textit{(Corresponding author: Peng Hu)}}
\thanks{Yongxiang Li, Yuan Sun, Yang Qin, Xi Peng and Peng Hu are with the College of Computer Science, Sichuan University, Chengdu 610065, China. (email: rhythmli.scu@gmail.com; penghu.ml@gmail.com).}
\thanks{Dezhong Peng is with the College of Computer Science, Sichuan University, Chengdu 610065, China, and also with the Sichuan National Innovation New Vision UHD Video Technology Co., Ltd, Chengdu 610095, China.}}

\markboth{Journal of \LaTeX\ Class Files,~Vol.~14, No.~8, August~2024}%
{Shell \MakeLowercase{\textit{et al.}}: A Sample Article Using IEEEtran.cls for IEEE Journals}

\markboth{IEEE TRANSACTIONS ON INFORMATION FORENSICS AND SECURITY, 2025}
{Shell \MakeLowercase{\textit{et al.}}: Bare Demo of IEEEtran.cls for IEEE Journals}
\maketitle

\begin{abstract}
Unsupervised visible-infrared person re-identification (UVI-ReID) aims at retrieving pedestrian images of the same individual across distinct modalities, presenting challenges due to the inherent heterogeneity gap and the absence of cost-prohibitive annotations. Although existing methods employ self-training with clustering-generated pseudo-labels to bridge this gap, they always implicitly assume that these pseudo-labels are predicted correctly. In practice, however, this presumption is impossible to satisfy due to the difficulty of training a perfect model let alone without any ground truths, resulting in pseudo-labeling errors. Based on the observation, this study introduces a new learning paradigm for UVI-ReID considering Pseudo-Label Noise (PLN), which encompasses three challenges: noise overfitting, error accumulation, and noisy cluster correspondence. To conquer these challenges, we propose a novel robust duality learning framework (RoDE) for UVI-ReID to mitigate the adverse impact of noisy pseudo-labels. Specifically, for noise overfitting, we propose a novel Robust Adaptive Learning mechanism (RAL) to dynamically prioritize clean samples while deprioritizing noisy ones, thus avoiding overemphasizing noise. To circumvent error accumulation of self-training, where the model tends to confirm its mistakes, RoDE alternately trains dual distinct models using pseudo-labels predicted by their counterparts, thereby maintaining diversity and avoiding collapse into noise. However, this will lead to cross-cluster misalignment between the two distinct models, not to mention the misalignment between different modalities, resulting in dual noisy cluster correspondence and thus difficult to optimize. To address this issue, a Cluster Consistency Matching mechanism (CCM) is presented to ensure reliable alignment across distinct modalities as well as across different models by leveraging cross-cluster similarities. Extensive experiments on three benchmark datasets demonstrate the effectiveness of the proposed RoDE.
\end{abstract}

\begin{IEEEkeywords}
Unsupervised VI-ReID; Pseudo-Label Noise; Noise Correspondence; Cluster Consistency
\end{IEEEkeywords}

\IEEEpeerreviewmaketitle

\section{Introduction}
\IEEEPARstart{V}{}isible-Infrared Person Re-Identification (VI-ReID) seeks to match pedestrians of the same identity across visible and infrared modalities, serving critical roles in various scenarios~\cite{cui2024dma,wu2023unsupervised,pang2023cross,yang2023towards} such as military surveillance, and intelligent security. This technology effectively enhances the precise identification and response capabilities in these fields, ensuring the overall security and stability of society. 

The major challenge in VI-ReID is learning modality-invariant representations to bridge the significant heterogeneity gap between visible and infrared data. To this end, numerous VI-ReID methods are proposed to project different modalities into a latent common space, which could be roughly categorized into supervised VI-ReID (SVI-ReID) and unsupervised VI-ReID (UVI-ReID). Specifically, SVI-ReID methods exploit identification labels to learn the semantic consistency across distinct modalities but are impractical due to the high cost of collecting a large amount of well-labeled multimodal data. Conversely, UVI-ReID methods circumvent this limitation, making them more practical~\cite{yang2022augmented}, but facing increased difficulty in deriving cross-modal consistency from unlabeled data.

\begin{figure}[t]
    \centering
    
    \begin{subfigure}[c]{0.49\linewidth}
        \includegraphics[width=\columnwidth]{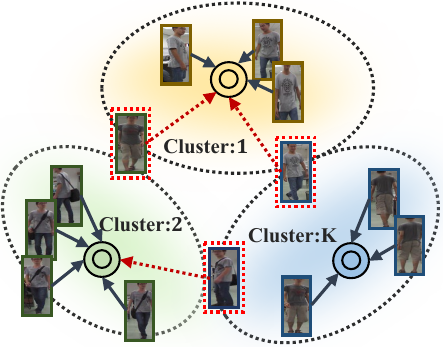}
        \caption{Noise Overfitting}
        \label{fig1a}
    \end{subfigure}
    \begin{subfigure}[c]{0.49\linewidth}
        \includegraphics[width=\columnwidth]{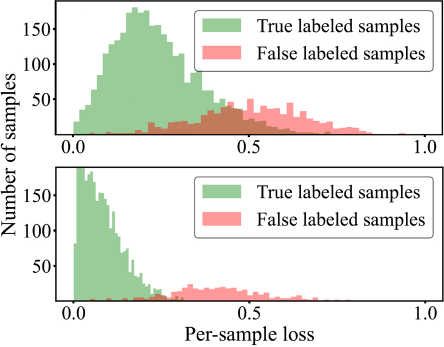}
        \caption{Error Accumulation}
        \label{fig1b}
    \end{subfigure}

    \begin{subfigure}[c]{0.96\linewidth}
        \includegraphics[width=\columnwidth]{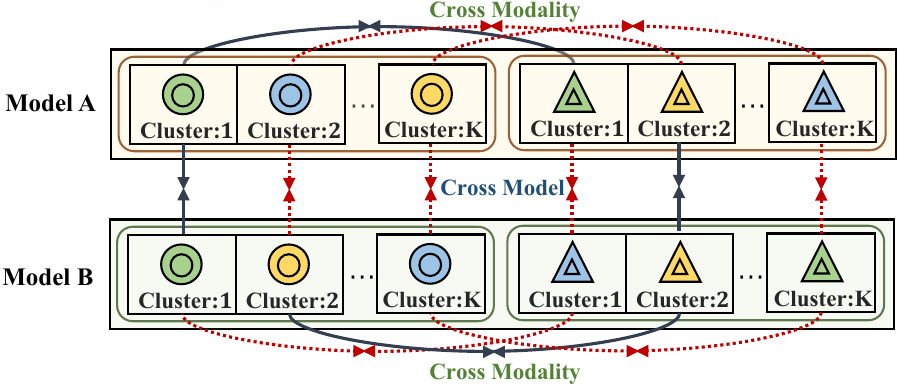}
        \caption{Noisy Cluster Correspondence}
        \label{fig1c}
    \end{subfigure}
    
    \begin{subfigure}[c]{0.96\linewidth}
        \includegraphics[width=\columnwidth]{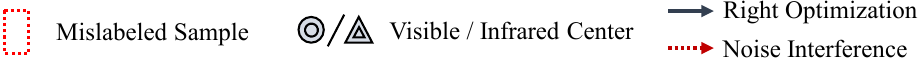}
    \end{subfigure}

    \caption{\small Pseudo-label noise issues in UVI-ReID. (a) In intra-modality, some sample features are close to the adjacent cluster center, leading to false pseudo-label assignments and noise overfitting. (b) Error accumulation for a single model (TOP) and dual models (BOTTOM) is depicted through the per-sample loss distribution on the infrared modality of RegDB dataset using the recent IMSL method~\cite{pang2024inter}. The dual models employ a cross-training strategy, using the pseudo-labels generated by one model as the ground truth for the other. During training, inevitable error annotations and cluster mismatches introduce significant noise. For a single model, noisy and clean samples intermingle due to severe error accumulation, as indicated by the overlapping color parts. In contrast, using dual models significantly mitigates this issue. (c) Semantic misalignment occurs across different clusters, including distinct models and modalities, which are regarded as noisy cluster correspondences.}
    \label{fig1}
\end{figure}

To tackle this pivotal challenge, some UVI-ReID methods employ clustering techniques to generate pseudo-labels for each modality~\cite{cheng2023unsupervised,shi2023dual}, thereby establishing cross-modal correspondences and learning modality-invariant representations. However, these methods ignore the problem of various types of noise interference caused by pseudo-labels, including noise overfitting, error accumulation, and noisy cluster correspondence, collectively referred to as Pseudo-Label Noise (PLN). These training noises often occur together and misguide the optimization of multimodal models, thereby leading to serious error accumulation and overfitting. To illustrate this problem, we depict it in~\Cref{fig1}, where it can be seen that PLN is a pervasive but neglected issue. Furthermore, while some researchers~\cite{yang2022learning} have explored twin noisy label problems (including noisy label and noisy correspondence) in SVI-ReID, they presume a reliable cross-modal consistency, which is absent in UVI-ReID due to the chaotic correspondence of modality-specific clusters. That is why we refer to it as noisy cluster correspondence rather than label noise, presenting a more daunting and complex challenge.

\begin{figure*}[!ht]
    \centering
    \begin{minipage}{0.8\linewidth}
        \centering
        \includegraphics[width=\linewidth]{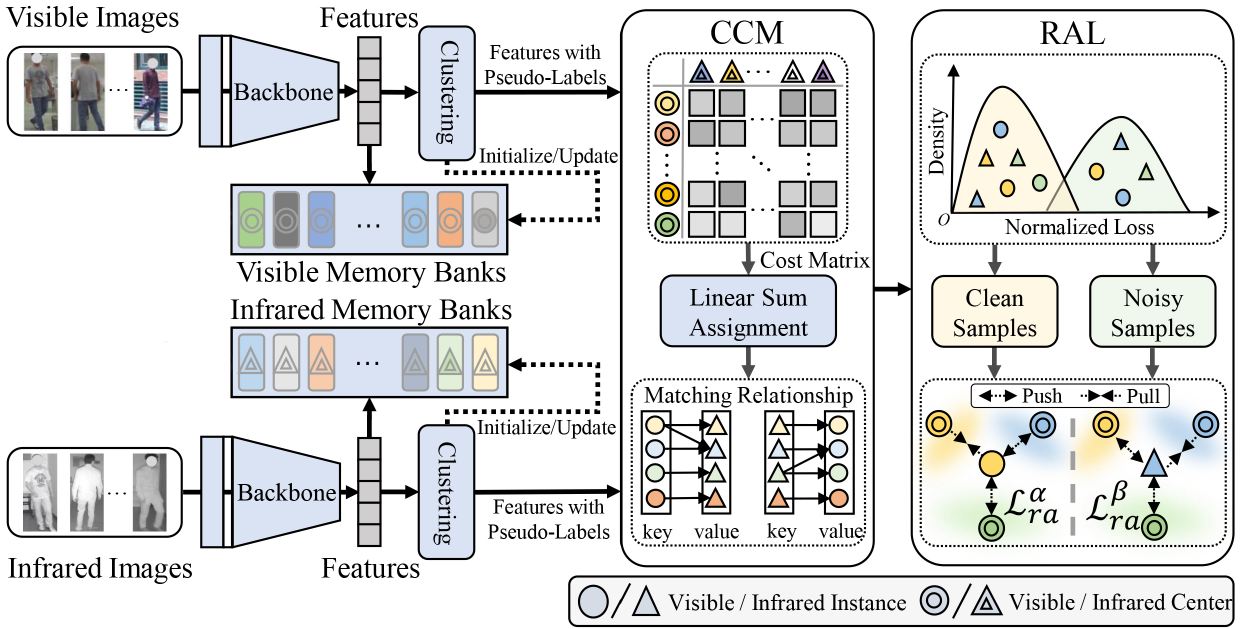}
    \end{minipage}
    \caption{\small The framework of the proposed RoDE. The model projects the visible and infrared images into the common space using the modality-specific networks $f^{\mathcal{P}}(\cdot; \Theta^{\mathcal{P}})$. CCM (See~\Cref{subsec:methodology5}) and RAL (See~\Cref{subsec:methodology3}) are used to alleviate noisy cluster correspondence and noisy overfitting. Specifically, cross-modal and cross-model CCM are utilized to establish the correspondence across different modalities and different models, respectively. Moreover, RAL divides the pseudo-labels into clean and noisy subsets, and adaptively adjusts the focus on them, thereby enhancing robustness against noisy overfitting.}
    \label{fig2}
\end{figure*}

To address the aforementioned challenges, we propose a novel unsupervised visible-infrared framework RoDE, to robustly learn from PLN using visible-infrared pedestrian images, as illustrated in~\Cref{fig2}. Specifically, RoDE incorporates the following tangible solutions: 1) To counter noise overfitting, we propose a Robust Adaptive Learning mechanism (RAL) that categorizes samples into clean and noisy subsets, and then dynamically prioritizes clean samples while deprioritizing noisy ones by using a robust loss function, thereby reducing the influence of mislabeled samples and mitigating overfitting noise. 2) To avoid error accumulation, we simultaneously train two different models using Robust Duality Learning (RDL), each using the predictions of the other model, thereby diversifying the supervision information. Thanks to this diversity, our RoDE could prevent each model from being overconfident about its own incorrect predictions, thus avoiding accumulating errors. 3) To tackle noisy cluster correspondence caused by the dual models, we present a novel Cluster Consistency Matching mechanism (CCM), which matches distinct clusters by utilizing the distances between different centers, producing more reliable correspondence. 

Unlike existing UVI-ReID methods that learn from pseudo-labels~\cite{li2024inter,cheng2023unsupervised,yang2022learning}, our RoDE addresses not only noise overfitting but also error accumulation. More specifically, most of these methods use a binary robust strategy~\cite{li2024inter,pang2023cross}, selectively focusing on confident samples and discarding all unreliable ones, which leads to information loss and performance degradation. In contrast, our RoDE reweights all samples adaptively, thereby avoiding the rough discard of data and reducing the adverse impact of noise. However, current robust methods focus solely on robust training techniques like sample selection and robust loss functions to alleviate noise, thereby becoming overconfident in their predictions even when incorrect, i.e. error accumulation. Our RoDE counters this issue by using two distinct models that alternately guide each other, diversifying supervision and preventing overconfidence, as shown in \Cref{fig1} (b). In addition, the intrinsic cross-modal and cross-model gaps lead to cross-cluster misalignment, referred to as dual noisy cluster correspondence. Intuitively, this noise would be more challenging than the single cross-cluster noise across different modalities in prior works~\cite{wu2023unsupervised}. To address this challenge, our RoDE matches distinct clusters across modalities and models, ensuring more reliable correspondence and enhancing overall robustness.

Our main contributions can be summarized as follows:
\begin{itemize}
    \renewcommand{\labelitemi}{$\bullet$}
    \item In this paper, we propose a novel UVI-ReID framework RoDE to robustly learn discriminative representations and establish cross-modal re-identification relationships in a latent common space, addressing noise overfitting, error accumulation, and noisy cluster correspondence simultaneously.
    \item To resist the interference of noisy overfitting, we design a novel RAL mechanism that utilizes a self-adaptive strategy and a demonstrably robust loss function to prioritize clean samples, thereby enhancing robustness.
    \item We present a RDL training pipeline that jointly trains two different models to prevent error accumulation in self-training. To meet UVI-ReID requirements, CCM mechanism is introduced to address noisy cluster correspondence, encompassing both cross-modal and cross-model scenarios.
    \item Extensive experiments on the SYSU-MM01, RegDB, and LLCM datasets highlight the superiority of our method and establish a powerful baseline for the UVI-ReID task.
\end{itemize}

\section{Related Works}
\label{sec:relatedworks}
\subsection{Supervised Visible-Infrared Person Re-Identification}
SVI-ReID is a subtask of cross model retrieval~\cite{sun2024dual}, which aims to match visible images of individuals with their infrared counterparts. To tackle cross-modal discrepancies, several supervised VI-ReID methods have been proposed to learn modality-invariant features~\cite{ye2019bi}. For example, HSME~\cite{hao2019hsme} uses a hypersphere manifold embedding with sphere softmax. MPANet leverages a joint modality and pattern alignment network to uncover cross-modal differences~\cite{wu2021discover}. TransVI employs a Transformer-based approach with a two-stream structure to capture modality-specific features and learn shared knowledge~\cite{chai2023dual}. Additionally,~\cite{wu2023style} introduces a style-agnostic framework that bridges modality gaps at both data and feature levels. However, these methods, while effective with ample cross-modal annotations, are limited in real-world scenarios due to their dependence on visible-infrared identity labels.

\subsection{Unsupervised Visible-Infrared Person Re-ID}
Recently, UVI-ReID has gained significant attention for its lower labeling costs and practical applications in night surveillance~\cite{pang2023cross,wu2023unsupervised,shi2024learning,shi2023dual,shi2025multi}. Unlike traditional methods, UVI-ReID cannot use cross-modal labeled pairs to learn modality-invariant features. Instead, a common approach is to generate pseudo-labels for images from each modality and use these pseudo-supervisions to learn a shared discriminative representation~\cite{wang2022optimal,bai2023rasa,cao2024empirical}. However, this strategy often requires additional RGB datasets for pre-training to acquire prior knowledge, which limits its practicality. Alternatively, some methods propose to generate pseudo-labels by exploring cluster-level relationships across different modalities through cross-modal memory aggregation, which can effectively capture the multimodal semantic consistency without any extra assistance~\cite{yang2022augmented,wu2023unsupervised,cao2022image}. Moreover, a series of methods pay attention to building cross-modal associations to embed domain information mutually, achieving remarkable performance~\cite{yang2023towards,cheng2023unsupervised}. However, these methods usually ignore the serious PLN problems during the training process or only notice a certain aspect of the impact of PLN, thereby misleading model optimization direction and degrading the performance.

\subsection{Robustly Learning with Noisy Labels}
The problem of noisy labels presents a significant challenge during training, potentially misdirecting the learning process~\cite{li2024romo,feng2023rono,sun2024robust}. To combat this negative influence, existing methods can be classified into three categories: sample selection, label correction, and noise regularization. Previous researches on sample selection aim to detect noisy labels by the natural resistance of neural networks to noise, often relying on batch statistics for robustness against label noise~\cite{huang2021learning}. Another research direction focuses on label correction, typically attempting to rectify sample labels using model predictions~\cite{yao2021jo}. Additionally, some studies emphasize noise regularization techniques such as mixup~\cite{zhang2017mixup}, or dedicated loss terms~\cite{hu2021learning}. Unsupervised regularization has also been proven to enhance the classification accuracy of neural networks when trained on noisy datasets. However, in UVI-ReID, we will face more challenging and complex problems caused by noisy labels, including noisy overfitting, error accumulation, and noisy cluster correspondence. These challenges necessitate novel strategies that can simultaneously address above issues to ensure robust and accurate model performance.

\section{Methodology}
\label{sec:methodology}

\subsection{Problem Statement and Notations}
\label{subsec:methodology1}
Let $\boldsymbol{\mathcal{X}} = \{\boldsymbol{\mathcal{X}}^{\mathcal{V}}, \boldsymbol{\mathcal{X}}^{\mathcal{I}}\}$ be the label-free visible-infrared training dataset, where $\boldsymbol{\mathcal{X}}^{\mathcal{V}} = \{\boldsymbol{x}_{1}^{\mathcal{V}}, \boldsymbol{x}_{2}^{\mathcal{V}}, \cdots, \boldsymbol{x}_{N^{\mathcal{V}}}^{\mathcal{V}}\}$ denotes the $N^{\mathcal{V}}$ visible images and $\boldsymbol{\mathcal{X}}^{\mathcal{I}} = \{\boldsymbol{x}_{1}^{\mathcal{I}}, \boldsymbol{x}_{2}^{\mathcal{I}}, \cdots, \boldsymbol{x}_{N^{\mathcal{I}}}^{\mathcal{I}}\}$ denotes the $N^{\mathcal{I}}$ infrared images. For convenience, $\boldsymbol{x}_{i}^{\mathcal{P}}$ is used to denote the $i$-th image in the $\mathcal{P} \in \{\mathcal{V}, \mathcal{I}\}$ modality, and $N^{\mathcal{P}}$ is the number of images in the $\mathcal{P}$ modality. UVI-ReID aims to learn common representations from the unlabeled and unaligned visible-infrared dataset $\boldsymbol{\mathcal{X}}$, thereby enabling the accurate retrieval of the most semantically relevant sample in another modality using the given query.

To achieve this, we first use two modality-specific nonlinear neural network projectors, $\{ f^{\mathcal{P}}(\cdot; \Theta^{\mathcal{P}}) \}_{\mathcal{P} \in \{\mathcal{V}, \mathcal{I}\}}$, to map images from each modality into an $L$-dimensional representation space, where $\Theta^{\mathcal{P}}$ represents the trainable parameters for the network of modality $\mathcal{P}$. In other words, each sample $\boldsymbol{x}_{i}^{\mathcal{P}}$ is projected as a feature vector $\boldsymbol{v}_{i}^{\mathcal{P}} \in \mathbb{R}^{1 \times L}$ by
\begin{equation}
    \begin{aligned}
        \boldsymbol{v}_{i}^{\mathcal{P}}=f^{\mathcal{P}}(\boldsymbol{x}_{i}^{\mathcal{P}}; \Theta^{\mathcal{P}}).
    \end{aligned}
    \label{eq1}
\end{equation}

To mine semantic information from unlabeled data, we follow prior works~\cite{wu2023unsupervised,yang2023towards} and employ classic clustering methods such as DBSCAN~\cite{ester1996density} to cluster the projected representations for each modality, assigning modality-specific pseudo-labels to each sample based on its nearest cluster center. Unfortunately, due to the lack of correspondence between modalities, establishing semantic associations between different modalities is challenging, which impedes the achievement of UVI-ReID. To address this issue, pseudo-labels across different modalities should be matched based on the distances between the cluster centers to mitigate the modality gap. Additionally, clustering to obtain pseudo-labels at each epoch introduces considerable variability for identical instances. To maintain the stability and reliability of pseudo-labels, we construct memory banks $\boldsymbol{\mathcal{M}}^{\mathcal{P}}$ to store all cluster centers, which are iteratively updated with newly generated cluster centers. This mechanism prevents the model from frequently reassigning different pseudo-labels to the same individuals across epochs, which could introduce confusion and instability in the training process.
\begin{equation}
    \begin{aligned}
        \boldsymbol{\mathcal{M}}^{\mathcal{P}} = [\boldsymbol{m}_1^{\mathcal{P}}, \cdots, \boldsymbol{m}_{K^{\mathcal{P}}}^{\mathcal{P}}] \in \mathbb{R}^{K^{\mathcal{P}} \times L},
    \end{aligned}
    \label{eq2}
\end{equation}
where $\boldsymbol{m}_i^{\mathcal{P}}$ represents the center of the $i$-th cluster, and $K^{\mathcal{P}}$ denotes the count of clusters in modality $\mathcal{P}$. The memory banks are iteratively updated through momentum after each epoch, i.e.,
\begin{equation}
    \begin{aligned}
        \boldsymbol{m}_{j}^{\mathcal{P}} = \eta \boldsymbol{m}_{j}^{\mathcal{P}} + (1-\eta)\Bar{\boldsymbol{v}}_j^{\mathcal{P}},
    \end{aligned}
    \label{eq3}
\end{equation} 
where $\Bar{\boldsymbol{v}}_{j}^{\mathcal{P}}$ is the mean feature in the $j$-th class, and $\eta \in \left[0, 1\right]$ is the memory updating rate. Notably, each center represents the mean feature of all samples within the same cluster, allowing the memory banks to semantically distinguish the samples based on their distances from the cluster centers in the feature space.

To learn representations that are both discriminative and modality-invariant under the supervision of pseudo-labels, the standard Cross-Entropy loss (CE) could be utilized to maximize the intra-modal conditional probabilities $p(\widetilde{y}_{i}^{\mathcal{P}} | \boldsymbol{x}_{i}^{\mathcal{P}})$ and the inter-modal conditional probabilities $p(\widetilde{y}_{i}^{\mathcal{Q}} | \boldsymbol{x}_{i}^{\mathcal{P}})$ as follows:
\begin{equation}
	\begin{aligned}
		\mathcal{L}_{ce} = - \sum_{\mathcal{P} \in \{\mathcal{V},\mathcal{I}\}} \sum_{i=1}^{N_B} & \Big( \widetilde{y}_{i}^{\mathcal{P}} \log \left( p(\widetilde{y}_{i}^{\mathcal{P}} | \boldsymbol{x}_{i}^{\mathcal{P}}) \right)  \\ 
          + & \widetilde{y}_{i}^{\mathcal{Q}} \log \left(p(\widetilde{y}_{i}^{\mathcal{Q}} | \boldsymbol{x}_{i}^{\mathcal{P}}) \right) \Big), 
	\end{aligned}
    \label{eq4}
\end{equation}
where $\boldsymbol{x}_{i}^{\mathcal{P}}$ belongs to the clustering center $\widetilde{y}_{i}^{\mathcal{P}}$ (i.e., pseudo-label). $\mathcal{Q} \in \{\mathcal{V}, \mathcal{I}\}$ and $\mathcal{Q} \ne \mathcal{P}$, $N_B$ is the batch size, and $p(\widetilde{y}_{i}^{\mathcal{P}} | \boldsymbol{x}_{i}^{\mathcal{P}})$ and $p(\widetilde{y}_{i}^{\mathcal{Q}} | \boldsymbol{x}_{i}^{\mathcal{P}})$ are calculated by:
\begin{equation}
	\begin{aligned}
        p(\widetilde{y}_{i}^{\mathcal{P}} | \boldsymbol{x}_{i}^{\mathcal{P}}) = \frac{\exp ((\boldsymbol{m}_{\widetilde{y}_{i}^{\mathcal{P}}}^{\mathcal{P}})^T \cdot \boldsymbol{v}_{i}^{\mathcal{P}} / \tau)}{\sum_{k=1}^{K^{\mathcal{P}}} \exp \left((\boldsymbol{m}_{k}^{\mathcal{P}})^T \cdot \boldsymbol{v}_{i}^{\mathcal{P}} / \tau \right) },
	\end{aligned}
    \label{eq5}
\end{equation}
\begin{equation}
	\begin{aligned}
        p(\widetilde{y}_{i}^{\mathcal{Q}} | \boldsymbol{x}_{i}^{\mathcal{P}}) = \frac{\exp ((\boldsymbol{m}_{\widetilde{y}_{i}^{\mathcal{Q}}}^{\mathcal{P}})^T \cdot \boldsymbol{v}_{i}^{\mathcal{P}} / \tau)}{\sum_{k=1}^{K^{\mathcal{P}}} \exp \left((\boldsymbol{m}_{k}^{\mathcal{P}})^T \cdot \boldsymbol{v}_{i}^{\mathcal{P}} / \tau \right) },
	\end{aligned}
    \label{eq6}
\end{equation}
where $\tau$ is a temperature parameter. Although existing methods achieve promising performance by using CE, they implicitly assume that the pseudo-labels are accurately annotated. Unfortunately, it is difficult or even possible to label the unlabeled data accurately, which inevitably introduces noise into pseudo-labels, leading to noise overfitting and error accumulation. This is also demonstrated in~\Cref{tab3}, where the noise in pseudo-labels disrupts the cross-modal association learning, seriously affecting re-identification performance.

\subsection{Robust Adaptive Learning Mechanism}
\label{subsec:methodology2}
To tackle the noise overfitting issue in UVI-ReID, most existing UVI-ReID approaches mitigate noise overfitting by employing binary robust strategies that selectively emphasize confident samples while disregarding unreliable ones~\cite{li2024inter,pang2023cross}. Although these methods reduce the impact of noise, their unreliable partitioning inadvertently leads to the loss of information from false negative samples, resulting in suboptimal performance. In contrast, we introduce the Robust Adaptive Learning (RAL) mechanism, which adaptively ensures a selective emphasis, effectively mitigating noise without sacrificing valuable information in unreliable samples. To be specific, to alleviate excessive optimization of samples with low reliability, we first design a robust loss function, $\mathcal{L}_{ra}$, explicitly to resist noise interference, which is defined as:
\begin{equation}
    \begin{aligned}
        \mathcal{L}_{ra} = - \sum_{\mathcal{P} \in \{\mathcal{V},\mathcal{I}\}} \sum_{i=1}^{N^{\mathcal{P}}} p^{\gamma_i} \left(\widetilde{y}_{i}^{\mathcal{P}} | \boldsymbol{x}_{i}^{\mathcal{P}}\right),
    \end{aligned}
    \label{eq7}
\end{equation}
where $\gamma_i \in \left(0, 1\right]$ is used to adjust the strength of optimization for each sample. Moreover, we provide the property of $\mathcal{L}_{ra}$ to better understand its robustness against noisy labels:

\textbf{\textit{Property 1:}} For any input ($\boldsymbol{x}_{i}^{\mathcal{P}}, \widetilde{y}_{i}^{\mathcal{P}}$) and $\gamma_i \in \left(0, 1\right]$,  the loss function $\mathcal{L}_{ra}$ exhibits the following behaviors:

\begin{enumerate}
    \item As $\gamma_i$ approaches to 0, $\mathcal{L}_{ra}$ gradually behaves like the CE loss.
    \item As $\gamma_i$ approaches to 1, $\mathcal{L}_{ra}$ tends to optimize equally for all samples.
\end{enumerate}

According to Property 1, one could infer that $\mathcal{L}_{ra}$ effectively reduces the focus on mislabeled samples, alleviating the overfitting issue caused by pseudo-labeling noise. Additionally, $\mathcal{L}_{ra}$ does not treat all samples equally, thereby mitigating the underfitting issue. Consequently, it improves performance while maintaining robustness against noise by appropriately attending to challenging samples. Meanwhile, the preference towards robustness and strong optimization is regulated by $\gamma_i$. Note that the detailed proofs for Property 1 are available in the Appendix.

However, due to the varying requirements of different samples for optimization, it is almost impossible to tune the parameter $\gamma_i$ for each sample manually, while using a single unified parameter for all samples would lead to suboptimal performance. To be ideal, the model should strongly optimize for confident samples (i.e., $\gamma_i \rightarrow 0$) while remaining robust to noisy labeled samples (i.e., $\gamma_i \rightarrow 1$). To this end, we present an adaptive method to determine the appropriate $\gamma_i$ for each sample, thereby avoiding the above dilemma. Specifically, we first define the per-sample loss as $\boldsymbol{\ell}^{\mathcal{P}}=\{\ell_{i}^{\mathcal{P}}\}_{i=1}^{N^{\mathcal{P}}}$ to measure the difference between predictions and pseudo-labels, where $\ell_{i}^{\mathcal{P}}$ is calculated by $\mathcal{L}_{i}^{\mathcal{P}}$ (See~\Cref{eq18}). In other words, this difference could reflect the reliability of the pseudo-labels. 
Therefore, based on the per-sample losses, we can use a two-component Gaussian Mixture Model (GMM) to classify samples with the lower mean as the clean set and those with the higher mean as the noisy set.
\begin{equation}
	\begin{aligned}
		p(\boldsymbol{\ell}^{\mathcal{P}}|\Theta^{G}) = \sum_{k=1}^{2} \delta_{k} \phi(\boldsymbol{\ell}^{\mathcal{P}}|k),
	\end{aligned}
    \label{eq8}
\end{equation}
where $\delta_{k}$ and $\phi(\boldsymbol{\ell}^{\mathcal{P}}|k)$ are the mixture coefficient and the probability density of the $k$-th component respectively, and $\Theta^{G}$ is the parameter of GMM. Although GMM is employed in our approach to distinguish clean and noisy samples, it is important to emphasize that GMM serves as one of several possible tools for data partitioning. Other clustering methods, such as K-Means or Beta Mixture Models (BMM), could also be utilized in place of GMM without compromising the fundamental contributions of our method. In addition, we compute the posterior probability $w_i=p(k)p(\ell_i^{\mathcal{P}}|k)/p(\ell_i^{\mathcal{P}})$ as the probability that the $i$-th sample belongs to the clean set. Based on the aforementioned discussion, it is optimal to assign small $\gamma_{i}$ to reliable samples while large $\gamma_{i}$ to noisy ones. To achieve this, we employ a sharpening strategy to calculate $\gamma_{i}$ adaptively as follows: 
\begin{equation}
    \begin{aligned}
        \gamma_{i}=\log \left((1-w_{i})^{0.25}/\mu+1\right),
    \end{aligned}
    \label{eq9}
\end{equation}
where $\mu$ is a scale parameter. Consequently, RAL can not only mitigate the detrimental effects of noise overfitting but also preserve useful information that might otherwise be lost.

\subsection{Robust Duality Learning Pipeline}
\label{subsec:methodology3}
\begin{figure*}[!ht]
    \centering
    \begin{minipage}{0.8\linewidth}
        \centering
        \includegraphics[width=\linewidth]{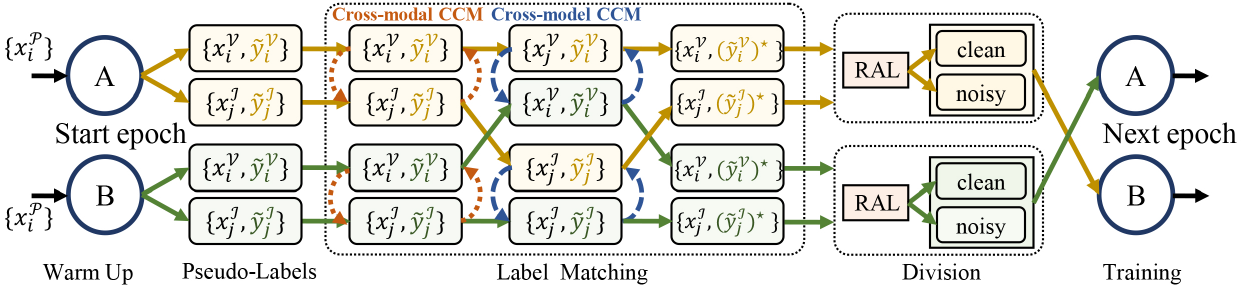}
    \end{minipage}
    
    \caption{\small The training pipeline of the proposed RoDE. RoDE consists of two individual models $A$ and $B$, which are trained collaboratively by exchanging their pseudo supervisions. Before training, RoDE pre-warms up the models $A$ and $B$ individually by predicting pseudo-labels and self-training. After warming up, the two models are co-trained with CCM and RAL.
    }
    \label{fig3}
\end{figure*}

To mitigate the interference of the noise in pseudo-labels, current methods predominantly focus only on robust training techniques~\cite{li2024inter,cheng2023unsupervised,yang2022learning}. However, these methods often neglect the issue of overconfidence in predictions even for incorrect ones, which leads to error accumulation as shown in \Cref{fig1} (b). To address this problem, we propose a Robust Duality Learning (RDL) pipeline to prevent the model from being overconfident about its own incorrect predictions, which alternately trains two individual networks with the same architecture but different initializations (denoted as $A=\{f^{\mathcal{P}}_{A}(\cdot;\Theta_{A}^{\mathcal{P}}), \boldsymbol{\mathcal{M}}^{\mathcal{P}}_{A}\}$ and $B=\{f^{\mathcal{P}}_{B}(\cdot;\Theta_{B}^{\mathcal{P}}), \boldsymbol{\mathcal{M}}^{\mathcal{P}}_{B}\}$), as depicted in~\Cref{fig3}. 

Specifically, in each epoch, models $A$ and $B$ first generate their own pseudo-labels $A:(\widetilde{y}_{i}^{\mathcal{P}} |_{A},\widetilde{y}_{i}^{\mathcal{Q}}|_{A})$ and $B:(\widetilde{y}_{i}^{\mathcal{P}}|_{B},\widetilde{y}_{i}^{\mathcal{Q}}|_{B})$ through a clustering method (e.g., DBSCAN~\cite{ester1996density}). Subsequently, model $A$ leverages the pseudo-labels generated by model $B$ for optimization, and vice versa. This mutual learning process enhances the diversity of learning and reduces the overconfidence to incorrect pseudo-labels in the specific model. However, the lack of pre-given correspondences can lead to unavoidable cluster mismatches across various modalities and models. This mismatch would seriously disrupt the optimization direction in the alternating learning process, resulting in poor performance. To establish cluster consistency, we propose the Cross-Cluster Matching (CCM) (see~\Cref{subsec:methodology4}), which aligns the two sets of clusters by utilizing correlations in both centers and identities, thereby producing the pseudo-label with reliable correspondences $A:((\widetilde{y}_{i}^{\mathcal{P}} |_{A})^{\star},(\widetilde{y}_{i}^{\mathcal{Q}}|_{A})^{\star})$ and $B:((\widetilde{y}_{i}^{\mathcal{P}}|_{B})^{\star},(\widetilde{y}_{i}^{\mathcal{Q}}|_{B})^{\star})$. Using the aligned labels, we optimize each model with RAL (see~\Cref{subsec:methodology2}) by minimizing both intra-modal loss $\mathcal{L}_{ra}^{\alpha}$ and inter-modal loss $\mathcal{L}_{ra}^{\beta}$. This dual loss minimization aims to reduce the impact of incorrectly labeled samples and enhance noise tolerance. Finally, the models are guided toward the correct optimization direction under cross supervision. To be specific, the objective functions $\mathcal{L}_{ra}^{\alpha}$ and $\mathcal{L}_{ra}^{\beta}$ of model $A$ could be rewritten as:
\begin{equation}
    \begin{aligned}
        \mathcal{L}_{ra}^{\alpha} = - \sum_{\mathcal{P} \in \{\mathcal{V},\mathcal{I}\}} \sum_{i=1}^{N^{\mathcal{P}}} p^{\gamma_i}\left( (\widetilde{y}_{i}^{\mathcal{P}}|_{B})^{\star} | \boldsymbol{x}_{i}^{\mathcal{P}}\right),
    \end{aligned}
    \label{eq10}
\end{equation}
\begin{equation}
    \begin{aligned}
        \mathcal{L}_{ra}^{\beta} = - \sum_{\mathcal{P} \in \{\mathcal{V},\mathcal{I}\}} \sum_{i=1}^{N^{\mathcal{P}}} p^{\gamma_i}\left( (\widetilde{y}_{i}^{\mathcal{Q}}|_{B})^{\star} | \boldsymbol{x}_{i}^{\mathcal{P}}\right),
    \end{aligned}
    \label{eq11}
\end{equation}
and the objective functions of model $B$ could be
\begin{equation}
    \begin{aligned}
        \mathcal{L}_{ra}^{\alpha} = - \sum_{\mathcal{P} \in \{\mathcal{V},\mathcal{I}\}} \sum_{i=1}^{N^{\mathcal{P}}} p^{\gamma_i}\left( (\widetilde{y}_{i}^{\mathcal{P}}|_{A})^{\star} | \boldsymbol{x}_{i}^{\mathcal{P}}\right),
    \end{aligned}
    \label{eq12}
\end{equation}
\begin{equation}
    \begin{aligned}
        \mathcal{L}_{ra}^{\beta} = - \sum_{\mathcal{P} \in \{\mathcal{V},\mathcal{I}\}} \sum_{i=1}^{N^{\mathcal{P}}} p^{\gamma_i}\left( (\widetilde{y}_{i}^{\mathcal{Q}}|_{A})^{\star} | \boldsymbol{x}_{i}^{\mathcal{P}}\right).
    \end{aligned}
    \label{eq13}
\end{equation}

\subsection{Cluster Consistency Matching}
\label{subsec:methodology4}
\begin{figure}[t]
    \centering
    \includegraphics[width=\columnwidth]{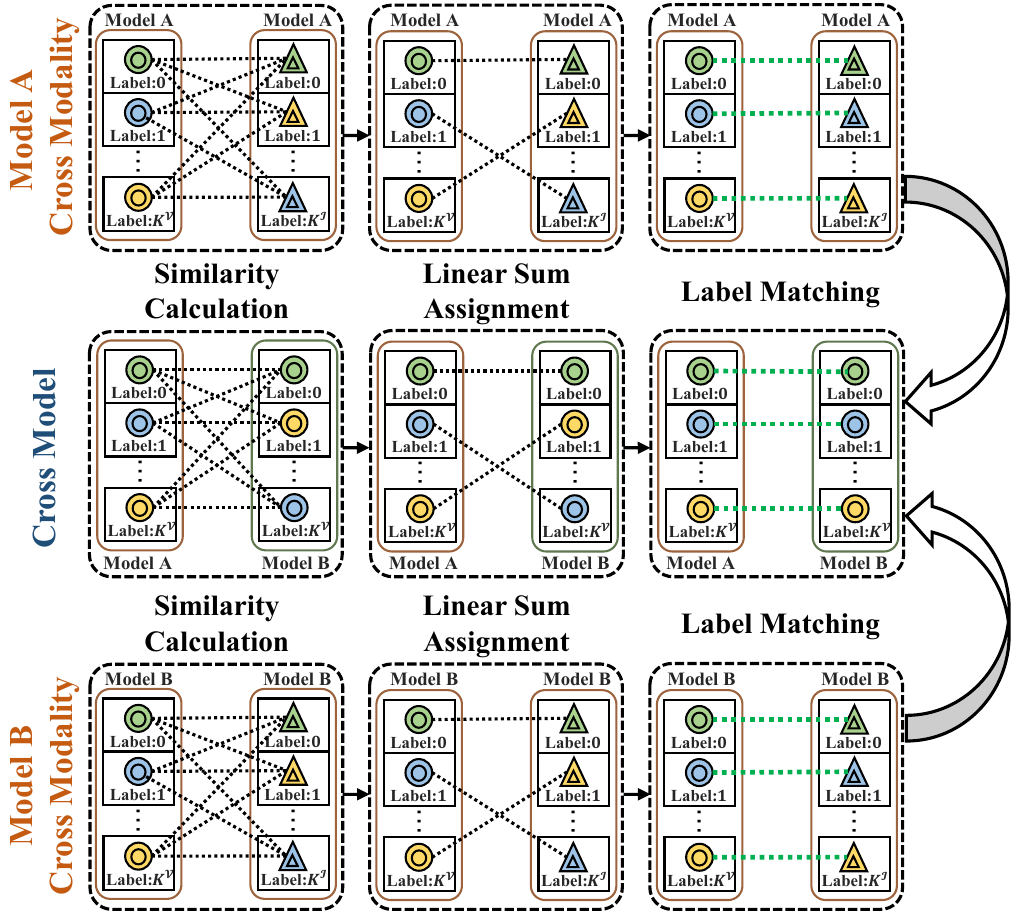}
    \caption{\small The solution of cluster inconsistency issue. \includegraphics[width=0.15in]{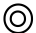}
    and \includegraphics[width=0.15in]{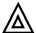} represent visible and infrared modality centers respectively. The \textit{\textcolor[rgb]{0.042,0.699,0.339}{green} dotted lines} denote correct matches after CCM.}
    \label{fig4}
\end{figure}

In the RDL pipeline, inherent discrepancies between cross-modal and cross-model elements lead to cross-cluster misalignment, referred to as dual noisy cluster correspondence. This form of noise presents a more significant challenge compared to the single cross-cluster noise encountered in previous studies on different modalities~\cite{wu2023unsupervised}. Dual noisy cluster correspondences exacerbate mismatching issues, leading to unstable training and even failure to converge. This significantly hinders the learning of association information between cross-modal and cross-model elements. To overcome this issue, we introduce the CCM mechanism, which aims to correlate the intrinsic characteristics of each cluster, thus aligning cluster centers across different modalities or models. Additionally, CCM accounts for the complex interactions between clusters across multiple modalities and models, providing a more comprehensive solution, as illustrated in~\Cref{fig4}.

In brief, the target of CCM could be formulated as a binary linear programming problem, which aims to match clusters with similar features. Specifically, we assume two sets of clustering center groups denoted as $\boldsymbol{C}^{P}=\{\boldsymbol{c}_{i}^{P}\}_{1 \le i \le N^{P}}$ and $\boldsymbol{C}^{Q}=\{\boldsymbol{c}_{j}^{Q}\}_{1 \le j \le N^{Q}}$, where $\boldsymbol{c}_{i}^{P}$ and $\boldsymbol{c}_{j}^{Q}$ represent the $i$-th and $j$-th cluster centers for the modalities or models $P$ and $Q$ respectively. $N^{P}$ and $N^{Q}$ are the numbers of clusters for $P$ and $Q$, respectively. Based on this, we design a cost matrix $\boldsymbol{S}=\{S_{ij}\}_{1 \le i \le N^{P}, 1 \le j \le N^{Q}}$, where $S_{ij}$ satisfies:
\begin{equation}
    \begin{aligned}
        S_{ij} = \exp\left(1 - \frac{\boldsymbol{c}_i^{P}}{{\| \boldsymbol{c}_i^{P} \|}}  \left(\frac{\boldsymbol{c}_j^{Q}}{\|\boldsymbol{c}_j^{Q}\|}\right)^T\right).
    \end{aligned}
    \label{eq14}
\end{equation}

Therefore, the objective function can be formulated as:
\begin{equation}
    \begin{aligned}
        \underset{\boldsymbol{M}} {\operatorname{arg\,min} \,} &\boldsymbol{S}^{T}\boldsymbol{M}, \\
        s.t. \;\; \boldsymbol{M}\boldsymbol{1}=\boldsymbol{1},~\boldsymbol{M}^T\boldsymbol{1}=\boldsymbol{1}&~\text{and}~\forall M_{ij} \in \{0,1\},
    \end{aligned}
    \label{eq15}
\end{equation}
where $\boldsymbol{1}$ is a column vector consisting entirely of ones, and the $\boldsymbol{M}$ serves as an indicator factor matrix, whose $(i, j)$-th element determines whether $\boldsymbol{c}_{i}^{P}$ and $\boldsymbol{c}_{j}^{Q}$ belong to the same class with $M_{ij}$ equaling 1, and 0 otherwise. Intuitively, many existing binary linear matching methods could be utilized to solve the problem of \Cref{eq15}, such as maximum weight matching, bipartite matching, and linear sum assignment. However, these methods would fail if $N^{P}$ and $N^{Q}$ are not equal, since they may leave many clusters unmatched in this case. To address this issue, we advocate for aligning unmatched clusters through multiple dynamic matches until all clusters have been progressively matched.

To mitigate inter-cluster misalignment, we perform consistency matching across different modalities and models, respectively. Specifically, we first align cross-modal clusters (i.e. \Cref{eq16}), ensuring that clusters from different modalities are harmonized. Then, we align clusters across distinct models (i.e. \Cref{eq17}) to ensure cross-model consistency and coherence. By maintaining consistency between the clusters produced by different models, we reduce the risk of error accumulation and enhance the overall robustness of the system. These two alignment strategies collectively improve the ability of model to accurately identify and match individuals across modalities.

\begin{equation}
    \text{Cross-modal Correspondence:} \left\{\begin{matrix}
      A:\mathcal{I} \leftrightarrow A:\mathcal{V} \\
      B:\mathcal{I} \leftrightarrow B:\mathcal{V}
    \end{matrix}\right.,
    \label{eq16}
\end{equation}

\begin{equation}
    \text{Cross-model Correspondence:} \left\{\begin{matrix}
      A:\mathcal{V} \leftrightarrow B:\mathcal{V} \\
      A:\mathcal{I} \leftrightarrow B:\mathcal{I}
    \end{matrix}\right..
    \label{eq17}
\end{equation}

Although the cross-modal/model gap challenges cluster alignment, two key strategies address it. First, common prior knowledge plays a crucial role. The backbones of both modalities/models are initialized with the same pre-trained weights, ensuring a shared feature representation that helps narrow the cross-modal gap from the start. Second, the robust loss in~\Cref{eq7} further strengthens alignment by guiding the model toward reliable correlations, prioritizing consistent clusters, and mitigating the impact of noisy correspondences.

\subsection{Training and Inference Strategy}
\label{subsec:methodology5}

\begin{algorithm}[t]
    \renewcommand{\algorithmicrequire}{\textbf{Input:}}
    \renewcommand{\algorithmicensure}{\textbf{Output:}}
    \caption{Optimization Procedure of RoDE.}
    \begin{algorithmic}[1]
    \REQUIRE Training dataset $\boldsymbol{\mathcal{X}}$, models $A$ and $B$, $\lambda$, $\tau$, $\eta$, epochs $N_e$, batch size $N_{b}$, and learning rate $lr$.
    \STATE Warm up $A$ and $B$ with~\Cref{eq4}, respectively;
    \FOR{$n_e$ $\rightarrow$ \{1, 2, $\cdots$, $N_{e}$\}}
        \STATE Calculate the features $\boldsymbol{v}_{i}^{\mathcal{P}} = f^{\mathcal{P}}(\boldsymbol{x}_{i}^{\mathcal{P}}; \Theta^{\mathcal{P}})$ for each sample $\boldsymbol{x}_{i}^{\mathcal{P}}$ using models $A$ and $B$;
        \STATE Perform modality-specific clustering on the features $\boldsymbol{v}_{i}^{\mathcal{P}}$ to acquire the centers for models $A$ and $B$;
        \IF{$n_e$ == 1}
            \STATE Initialize memory banks $\boldsymbol{\mathcal{M}}_{A}^{\mathcal{P}}$ and $\boldsymbol{\mathcal{M}}_{B}^{\mathcal{P}}$ using the clustering centers;
        \ELSE
            \STATE Update memory banks $\boldsymbol{\mathcal{M}}_{A}^{\mathcal{P}}$ and $\boldsymbol{\mathcal{M}}_{B}^{\mathcal{P}}$ with~\Cref{eq3};
            \STATE Align cluster centers across different modalities or models through~\Cref{eq16} and~\Cref{eq17}.
            \STATE Generate model-specific pseudo-labels $(\widetilde{y}_{i}^{\mathcal{P}}|_{A},\widetilde{y}_{i}^{\mathcal{Q}}|_{A})$ and $(\widetilde{y}_{i}^{\mathcal{P}}|_{B},\widetilde{y}_{i}^{\mathcal{Q}}|_{B})$;
            \STATE Obtain pseudo-labels with cross cluster consistency, i.e. $((\widetilde{y}_{i}^{\mathcal{P}}|_{A})^{\star},(\widetilde{y}_{i}^{\mathcal{Q}}|_{A})^{\star})$ and $((\widetilde{y}_{i}^{\mathcal{P}}|_{B})^{\star},(\widetilde{y}_{i}^{\mathcal{Q}}|_{B})^{\star})$;
            \REPEAT
                \STATE Randomly select $N_{b}$ samples;
                \STATE Update parameters $\Theta_{A}^{\mathcal{P}}$ with~\Cref{eq19} using the pseudo-labels $((\widetilde{y}_{i}^{\mathcal{P}}|_{B})^{\star},(\widetilde{y}_{i}^{\mathcal{Q}}|_{B})^{\star})$;
                \STATE Update parameters $\Theta_{B}^{\mathcal{P}}$ with~\Cref{eq19} using the pseudo-labels $((\widetilde{y}_{i}^{\mathcal{P}}|_{A})^{\star},(\widetilde{y}_{i}^{\mathcal{Q}}|_{A})^{\star})$;
            \UNTIL{\textit{All samples are selected;}}
        \ENDIF
    \ENDFOR
    \ENSURE Optimized parameters $\Theta_{A}^{\mathcal{P}}$ and $\Theta_{B}^{\mathcal{P}}$.
    \end{algorithmic}  
    \label{alg1}
\end{algorithm}

In a specific model $A$ or $B$, given an input image $\boldsymbol{x}_{i}^{\mathcal{P}}$, its training loss $\mathcal{L}_{i}^{\mathcal{P}}$ is actually a combination of intra-modal loss and inter-modal loss with a trade-off parameter $\lambda$. $\mathcal{L}_{i}^{\mathcal{P}}$ is defined as follow:
\begin{equation}
    \hspace{-4mm}
    \begin{aligned}
        \mathcal{L}_{i}^{\mathcal{P}} = -\lambda p^{\gamma_i}\left( (\widetilde{y}_{i}^{\mathcal{P}}|_{Z})^{\star} | \boldsymbol{x}_{i}^{\mathcal{P}}\right) - (1 - \lambda) p^{\gamma_i}\left( (\widetilde{y}_{i}^{\mathcal{Q}}|_{Z})^{\star} | \boldsymbol{x}_{i}^{\mathcal{P}}\right), 
    \end{aligned}
    \label{eq18}
\end{equation}
where $Z$ refers to the model $A$ or $B$, i.e., $Z \in \{A, B\}$. The overall loss $\mathcal{L}_{all}$ can be formulated as:
\begin{equation}
    \begin{aligned}
        \mathcal{L}_{all} & = \sum_{\mathcal{P} \in \{\mathcal{V},\mathcal{I}\}} \sum_{i=1}^{N^{\mathcal{P}}} \mathcal{L}_{i}^{\mathcal{P}} \\ 
        & = \lambda \underbrace{\sum_{\mathcal{P} \in \{\mathcal{V},\mathcal{I}\}} \sum_{i=1}^{N^{\mathcal{P}}} -  p^{\gamma_i}\left( (\widetilde{y}_{i}^{\mathcal{P}}|_{Z})^{\star} | \boldsymbol{x}_{i}^{\mathcal{P}}\right)}_{\mathcal{L}_{ra}^{\alpha}} \\ & + (1 - \lambda) \underbrace{\sum_{\mathcal{P} \in \{\mathcal{V},\mathcal{I}\}} \sum_{i=1}^{N^{\mathcal{P}}} -  p^{\gamma_i}\left( (\widetilde{y}_{i}^{\mathcal{Q}}|_{Z})^{\star} | \boldsymbol{x}_{i}^{\mathcal{P}}\right)}_{\mathcal{L}_{ra}^{\beta}}. 
    \end{aligned}
    \label{eq19}
\end{equation}

Overall, the training process alternates between two models initialized with distinct parameters by minimizing the objective $\mathcal{L}_{all}$. Notably, each model uses pseudo-labels generated by the other model for optimization. The detailed optimization procedure of RoDE is given in~\Cref{alg1}.

In the inference stage, we integrate the features of models $A$ and $B$ to obtain more comprehensive and robust representations, thus enhancing the representation capability and improving the performance. More specifically, given a query image $\boldsymbol{x}_{i}^{\mathcal{P}}$, its corresponding joint feature is:
\begin{equation}
    \begin{aligned}
        \boldsymbol{v}_{i}^{\mathcal{P}}=\frac{1}{2} \left(f_{A}^{\mathcal{P}}(\boldsymbol{x}_{i}^{\mathcal{P}}; \Theta_{A}^{\mathcal{P}}) + f_{B}^{\mathcal{P}}(\boldsymbol{x}_{i}^{\mathcal{P}}; \Theta_{B}^{\mathcal{P}})\right).
    \end{aligned}
    \label{eq20}
\end{equation}
Subsequently, we use the joint features $\boldsymbol{v}_{i}^{\mathcal{P}}$ to identify the pedestrian image with the highest cross-modal similarity, thereby obtaining the re-identification results for $\boldsymbol{x}_{i}^{\mathcal{P}}$.

\section{Experiments}
\label{sec:experiments}

\begin{table*}[!th]
    \small
    \centering
    \setlength{\tabcolsep}{2pt}
    \caption{\small Comparison with the recent methods on SYSU-MM01 and RegDB datasets. Rank-1 (\%), Rank-10 (\%), Rank-20 (\%), mAP (\%) and mINP (\%) are reported. The highest score is shown in \textbf{bold}, while the second highest score is \underline{underlined}.}
    \setlength{\tabcolsep}{1.75mm}{
    \begin{tabular*}{\textwidth}
    {@{\extracolsep{\fill}}c|l|l|ccc|ccc|ccc|ccc@{\extracolsep{\fill}}}
        \toprule[1.25 pt]
        \multirow{3}{*}{} & \multirow{3}{*}{Methods} & \multirow{3}{*}{Venue} & \multicolumn{6}{c|}{\makecell[c]{SYSU-MM01}} & \multicolumn{6}{c}{\makecell[c]{RegDB}} \\
        \cline{4-15}
         & & & \multicolumn{3}{c|}{\makecell[c]{All Search}} 
        & \multicolumn{3}{c|}{\makecell[c]{Indoor Search}} & \multicolumn{3}{c|}{\makecell[c]{V2T}} 
        & \multicolumn{3}{c}{\makecell[c]{T2V}} \\
        \cline{4-15}
            & & & Rank-1 & mAP & mINP & Rank-1 & mAP & mINP & Rank-1 & mAP & mINP & Rank-1 & mAP & mINP\\
        \midrule
        \multirow{9}{*}{\rotatebox{90}{SVI-ReID}} & AGW~\cite{ye2021deep}         & TPAMI'\textcolor{black}{21} & 47.50 & 47.65 & 35.30 & 54.17 & 62.97 & 59.23 & 70.05 & 66.37 & 50.19 & 70.49 & 65.90 & 51.24 \\
                                                  & CA~\cite{ye2021channel}       & ICCV'\textcolor{black}{21} & 69.88 & 66.89 & 53.61 & 76.26 & 80.37 & 76.79 & 85.03 & 79.14 & 65.33 & 84.75 & 77.82 & 61.56 \\
                                                  & DART~\cite{yang2022learning}  & CVPR'\textcolor{black}{22} & 68.72 & 66.29 & 53.26 & 72.52 & 78.17 & 74.94 & 83.60 & 75.67 & - & 81.97 & 75.13 & - \\
                                                  & SPOT~\cite{chen2022structure} & TIP'\textcolor{black}{22} & 65.34 & 62.25 & 48.86 & 69.42 & 74.63 & 70.48 & 80.35 & 72.46 & 56.19 & 79.37 & 72.26 & 56.06 \\
                                                  & CMTR~\cite{liang2023cross}    & TMM'\textcolor{black}{23} & 65.45 & 62.90 & - & 71.46 & 76.67 & - & 88.11 & 81.66 & - & 84.92 & 80.79 & - \\
                                                  & PMT~\cite{lu2023learning}     & AAAI'\textcolor{black}{23} & 67.53 & 64.98 & 51.86 & 71.66 & 76.52 & 72.74 & 84.83 & 76.55 & - & 84.16 & 75.13 & - \\
                                                  & TransVI~\cite{chai2023dual}   & TCSVT'\textcolor{black}{23} & 71.36 & 68.63 & - & 77.40 & 81.31 & - & 96.66 & 91.22 & - & 96.30 & 91.21 & - \\
                                                  & STAR~\cite{wu2023style}       & TMM'\textcolor{black}{23} & 76.07 & 72.73 & - & 83.47 & 85.76 & - & 94.09 & 88.75 & - & 93.30 & 88.20 & - \\ 
                                                  & DMA~\cite{cui2024dma}         & TIFS'\textcolor{black}{24} & 74.57 & 70.41 & 56.50 & 82.85 & 85.10 & - & 93.30 & 88.34 & - & 91.50 & 86.80 & - \\ 
        \midrule                                       
        \multirow{18}{*}{\rotatebox{90}{UVI-ReID}} 
                                                  & SPCL~\cite{ge2020self}        & NIPS'\textcolor{black}{20}  & 18.37 & 19.39 & 10.99 & 26.83 & 36.42 & 33.05 & 13.59 & 14.86 & 10.36 & 11.70 & 13.56 & 10.09 \\
                                                  & MMT~\cite{ge2020mutual}       & ICLR'\textcolor{black}{20}  & 21.47 & 21.53 & 11.50 & 22.79 & 31.50 & 27.66 & 25.68 & 26.51 & 19.56 & 25.59 & 18.66 & - \\
                                                  & IICS~\cite{xuan2021intra}     & CVPR'\textcolor{black}{21}  & 14.39 & 15.74 & 8.41 & 15.91 & 24.87 & 22.15 & 10.30 & 11.94 & 8.10 & 10.39 & 11.23 & 7.04 \\
                                                  & CAP~\cite{wang2021camera}     & AAAI'\textcolor{black}{21}  & 16.82 & 15.71 & 7.02 & 24.57 & 30.74 & 26.15 & 84.83 & 76.55 & - & 84.16 & 75.13 & - \\
                                                  & H2H~\cite{liang2021homogeneous}    & TIP'\textcolor{black}{21}  & 23.81 & 18.87 & - & - & - & - & 13.91 & 12.72 & - & 14.11 & 12.29 & - \\
                                                  & PPLR~\cite{cho2022part}            & CVPR'\textcolor{black}{22}  & 11.98 & 12.25 & 4.97 & 12.71 & 20.81 & 17.61 & 10.30 & 11.94 & 8.10 & 10.39 & 11.23 & 7.04 \\
                                                  & OTAL~\cite{wang2022optimal}        & ECCV'\textcolor{black}{22}  & 29.90 & 27.10 & - & 29.80 & 38.80 & - & 32.90 & 29.70 & - & 32.10 & 28.60 & - \\
                                                  & ADCA~\cite{yang2022augmented}      & MM'\textcolor{black}{22} & 45.51 & 42.73 & 28.29 & 50.60 & 59.11 & 55.17 & 67.20 & 64.05 & 52.67 & 68.48 & 63.81 & 49.62 \\
                                                  & DOTLA~\cite{cheng2023unsupervised} & MM'\textcolor{black}{23} & 50.36 & 47.36 & 32.40 & 53.47 & 61.73 & 57.35 & \underline{85.63} & 76.71 & \underline{61.58} & \underline{82.91} & 74.97 & \underline{58.60} \\
                                                  & CCLNet~\cite{chen2023unveiling}    & MM'\textcolor{black}{23} & 54.03 & 50.19 & - & 56.68 & 65.12 & - & 69.94 & 65.53 & - & 70.17 & 66.66 & - \\
                                                  & GUR~\cite{yang2023towards}         & ICCV'\textcolor{black}{23} & \underline{60.95} & \underline{56.99} & \underline{41.85} & \underline{64.22} & \underline{69.49} & \underline{64.81} & 73.91 & 70.23 & 58.88 & 75.00 & 69.94 & 56.21 \\
                                                  & PGM~\cite{wu2023unsupervised}      & CVPR'\textcolor{black}{23} & 57.27 & 51.78 & 34.96 & 56.23 & 62.74 & 58.13 & 69.48 & 65.41 & 38.72 & 69.85 & 65.17 & 58.47 \\
                                                  & DFC~\cite{si2023diversity}         & IPM'\textcolor{black}{23}    & 40.92 & 36.20 & - & 44.12 & 28.36 & - & 38.88 & 38.11 & - & - & - & - \\ 
                                                  & CHCR~\cite{pang2023cross}          & TCSVT'\textcolor{black}{23}  & 47.72 & 45.34 & - & - & - & - & 68.18 & 63.75 & - & 69.08 & 63.95 & - \\
                                                  & TAA~\cite{yang2023translation}     & TIP'\textcolor{black}{23}    & 48.77 & 42.43 & 25.37 & 50.12 & 56.02 & 49.96 & 62.23 & 56.00 & 41.51 & 63.79 & 56.53 & 38.99 \\ 
                                                  & IMSL~\cite{pang2024inter}          & TCSVT'\textcolor{black}{24}    & 57.96 & 53.93 & - & 58.30 & 64.31 & - & 70.08 & 66.30 & - & 70.67 & 66.35 & - \\
                                                  & SCA-RCP~\cite{li2024inter}         & TKDE'\textcolor{black}{24}    & 51.41 & 48.52 & 33.56 & 56.77 & 64.19 & 59.25 & 85.59 & \underline{78.12} & - & 82.41 & \underline{75.73} & - \\  
                                                  & RoDE        &    Ours    &  \textbf{62.88} & \textbf{57.91} & \textbf{43.04} & \textbf{64.53} & \textbf{70.42} & \textbf{66.04} & \textbf{88.77} & \textbf{78.98} & \textbf{67.99} & \textbf{85.78} & \textbf{78.43} & \textbf{62.34} \\
        \bottomrule[1.25 pt]
    \end{tabular*}}
    \label{tab1}
\end{table*}

\subsection{Datasets}
\label{subsec:experiment1}
We evaluate our proposed RoDE using three publicly available datasets:

\textbf{SYSU-MM01}~\cite{wu2017rgb} is a large-scale visible-infrared person re-identification dataset with four visible and two near-infrared cameras, covering both indoor and outdoor settings. The training set includes 22,257 visible images and 11,909 infrared images of 395 identities. For single-shot evaluation, the query and gallery sets comprise 3,803 infrared images and 301 randomly selected visible images from 96 identities.

\textbf{RegDB}~\cite{nguyen2017person} is a smaller dataset with two aligned cameras (one visible and one thermal). The similarity in body pose and capture distance between modalities reduces the challenge of visible-infrared re-identification. Each identity is represented by 10 visible and 10 infrared images.

\textbf{LLCM}~\cite{nguyen2017person} is a low-light multimodal dataset with 9 cameras in dim environments, including 46,767 bounding boxes across 1,064 identities. Both modalities suffer from issues like blurring and pose variation.

\begin{table}[!ht]
    \small
    \centering
    \setlength{\tabcolsep}{4pt}
    \caption{\small Comparison with the recent methods on LLCM dataset. Rank-1 (\%) and mAP (\%) are reported.}
    \setlength{\tabcolsep}{1.5mm}{
    \begin{tabular*}{\columnwidth}{@{\extracolsep{\fill}}c|l|l|cc|cc@{\extracolsep{\fill}}}
        \toprule[1.25 pt]
        & \multirow{2}{*}{Methods} & \multirow{2}{*}{Venue} & \multicolumn{2}{c|}{\makecell[c]{V2T}} & \multicolumn{2}{c}{\makecell[c]{T2V}} \\
        \cline{4-7}
                           & & & Rank-1 & mAP & Rank-1 & mAP \\
        \midrule
        \multirow{4}{*}{\rotatebox{90}{SVI-ReID}}   
                                                    & LBA~\cite{park2021learning}       & ICCV'\textcolor{black}{21} & 50.85 & 55.63 & 43.61 & 51.86 \\
                                                    & AGW~\cite{ye2021deep}  & TPAMI'\textcolor{black}{21} & 51.51 & 55.34 & 43.65 & 51.87 \\
                                                    & MMN~\cite{zhang2021towards} & MM'\textcolor{black}{21} & 59.97 & 62.75 & 52.53 & 58.99 \\
                                                    & DEEN~\cite{zhang2023diverse}    & CVPR'\textcolor{black}{23} & 62.57 & 65.81 & 54.92 & 62.95 \\
        \midrule
        \multirow{9}{*}{\rotatebox{90}{UVI-ReID}}  & CAP~\cite{wang2021camera}    & AAAI'\textcolor{black}{21} & 8.16 & 10.14 & 7.28 & 9.67 \\
                                                    & P2LR~\cite{han2022delving}     & AAAI'\textcolor{black}{22} & 16.38 & 19.84 & 14.85 & 17.15 \\
                                                    & OTLA~\cite{wang2022optimal}          & ECCV'\textcolor{black}{22} & 17.88 & 20.46 & 14.97 & 18.66 \\
                                                    & ADCA~\cite{yang2022augmented}     & MM'\textcolor{black}{22} & 23.57 & 28.25 & 16.16 & 21.48 \\
                                                    & GUR~\cite{yang2023towards}     & ICCV'\textcolor{black}{23} & \underline{31.47} & \underline{34.77} & \underline{29.68} & \underline{33.38} \\
                                                    & DOTLA~\cite{cheng2023unsupervised}     & MM'\textcolor{black}{23} & 27.14 & 26.26 & 23.52 & 27.48 \\
                                                    & IMSL~\cite{pang2024inter}     & TCSVT'\textcolor{black}{24} & 22.74 & 19.38 & 17.26 & 24.38 \\
                                                    & SCA-RCP~\cite{li2024inter}     & TKDE'\textcolor{black}{24} & 29.11 & 33.33 & 22.36 & 28.05 \\
                                                    & RoDE        & Ours & \textbf{35.13} & \textbf{37.44} & \textbf{32.73} & \textbf{36.64} \\
        \bottomrule[1.25 pt]
    \end{tabular*}}
    \label{tab2}
\end{table}

\subsection{Evaluation Metric}
\label{subsec:experiment2}
To ensure fair comparisons, we use established protocols to evaluate retrieval performance~\cite{yang2022learning}. These include Cumulative Matching Characteristic (CMC), mean Average Precision (mAP), and mean Inverse Negative Penalty (mINP). Following~\cite{wu2021discover}, we assess SYSU-MM01 dataset performance using both testing modes across 10 randomly selected gallery sets. For RegDB and LLCM, we consider two scenarios: Visible to Thermal (V2T) and Thermal to Visible (T2V). The training is performed entirely in an unsupervised manner, with identity labels used only during testing.

\subsection{Experimental Settings}
\label{subsec:experiment3}
The experiments and evaluations of RoDE are conducted on four NVIDIA Tesla V100 GPUs with Ubuntu 18.04.6 OS using PyTorch. We utilize AGW~\cite{ye2021deep} as the feature extractor for both visible and infrared modalities. All the input images are resized to 288 $\times$ 144 and then executed data augmentation, including random flipping, random erasing, and random cropping. In RoDE, the initial learning rate is set to 3.5 $\times$ 10$^{-4}$ and decays by a factor of 10 every 25 epochs. We train the model for a total of 50 epochs. The batch size is 32, with a memory updating rate $\eta$ of 0.15 and a temperature factor $\tau$ of 0.05. The trade-off parameter $\lambda$ is analyzed in~\Cref{subsec:experiment5}.

\subsection{Comparison with the State-of-the-art Methods}
\label{subsec:experiment4}
To evaluate the effectiveness of our RoDE, we compare it with 26 state-of-the-art methods across three benchmark datasets. These methods are grouped into two categories: 9 supervised VI-ReID (SVI-ReID) methods and 17 unsupervised VI-ReID (UVI-ReID) methods. From the results in~\Cref{tab1,tab2}, one can be drawn the following observations:
\begin{itemize}
    \item \textit{Comparison with UVI-ReID Methods:} Our RoDE achieves state-of-the-art performance on the three benchmark datasets in the unsupervised setting. To be specific, on the SYSU-MM01 dataset, RoDE achieves 62.88\% Rank-1, 57.91\% mAP, and 43.04\% mINP in the All Search mode, and 64.53\% Rank-1, 70.42\% mAP, and 66.04\% mINP in the Indoor-Search mode. On the RegDB dataset, RoDE shows significant advancements over the latest SCA-RCP~\cite{li2024inter}, with a notable Rank-1 improvement of 3.18\% in V2T and 3.37\% in T2V. Moreover, on the more challenging LLCM dataset, RoDE demonstrates both outstanding and promising performance. Compared to the second-best methods, GUR~\cite{yang2023towards}, it achieves a 3.66\% Rank-1 increase in the V2T setting and a 3.05\% Rank-1 increase in the T2V setting, respectively. Overall, these observations highlight the immense potential of RoDE, particularly in real-world scenarios that require high accuracy and involve challenging conditions.
    \item \textit{Comparison with Noisy-labels based UVI-ReID Methods:} Compared to existing methods, our RoDE systematically considered handling various noise issues raised by pseudo-labels. More importantly, RoDE reveals error accumulation in unsupervised cross-modal learning, which has been overlooked by previous noisy label learning based UVI-ReID methods such as CHCR~\cite{pang2023cross}, DOTLA~\cite{cheng2023unsupervised}, IMSL~\cite{pang2024inter}, and SCA-RCP~\cite{li2024inter}. From the results, one could find that our RoDE outperforms these methods in overall performance, demonstrating the effectiveness of RDL in tackling error accumulation.
    \item \textit{Comparison with SVI-ReID Methods:} Our RoDE achieves comparable performance to or even surpasses some supervised methods, especially on the RegDB dataset, which demonstrates the superiority of RoDE in effectively extracting discrimination from unlabeled and unaligned VI-ReID data. However, UVI-ReID still faces challenges in achieving a more precise cross-modal semantic understanding, indicating potential space for improvement.
\end{itemize}

\subsection{Parameter Analysis}
\label{subsec:experiment5}
\begin{figure}[ht]
    \centering    
    \begin{subfigure}[c]{0.49\linewidth}
        \centering
        \includegraphics[width=\columnwidth]{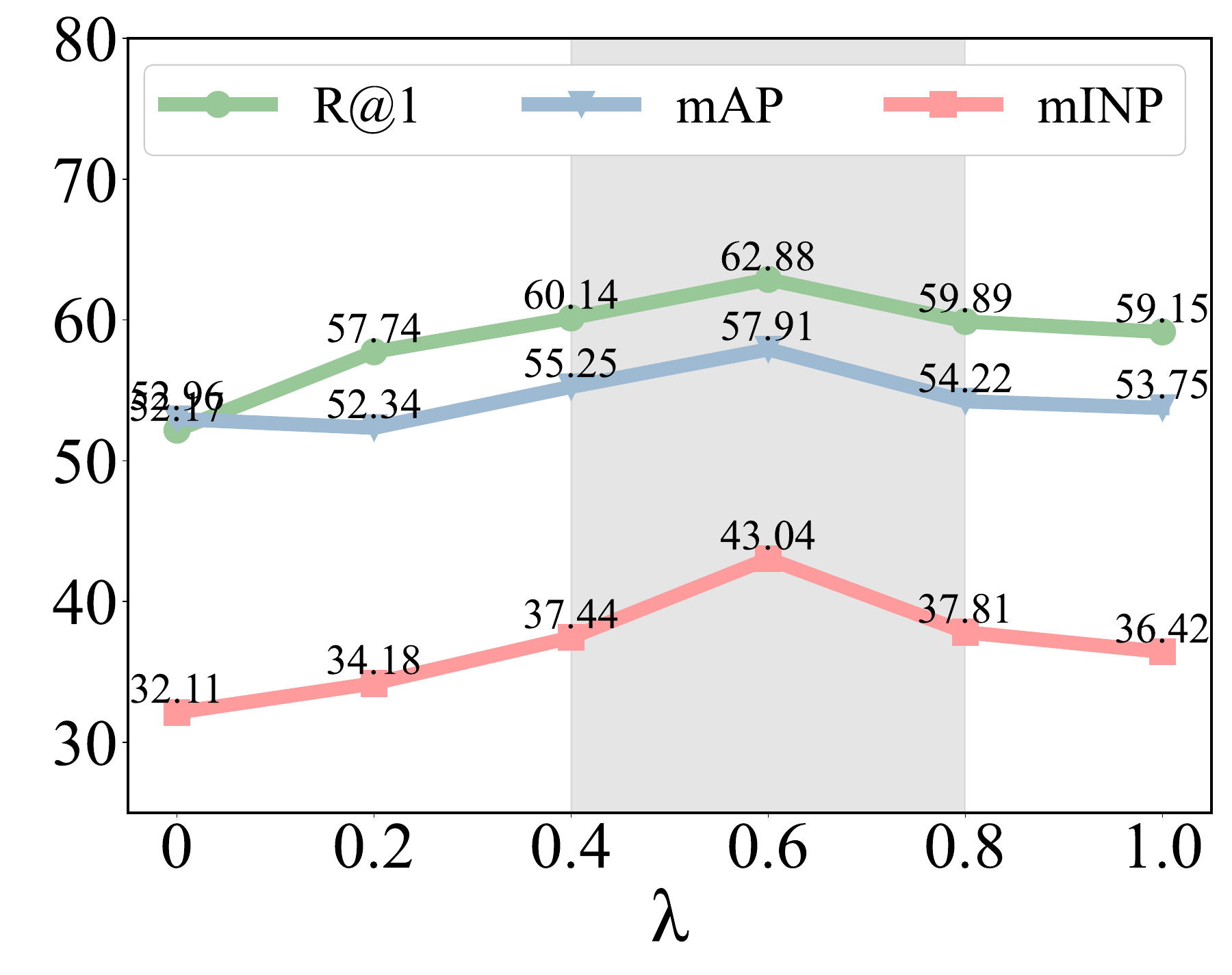}
        \caption*{(a) SYSU-MM01:All Search}
        \label{fig5a}
    \end{subfigure}  
    \begin{subfigure}[c]{0.49\linewidth}
        \centering
        \includegraphics[width=\columnwidth]{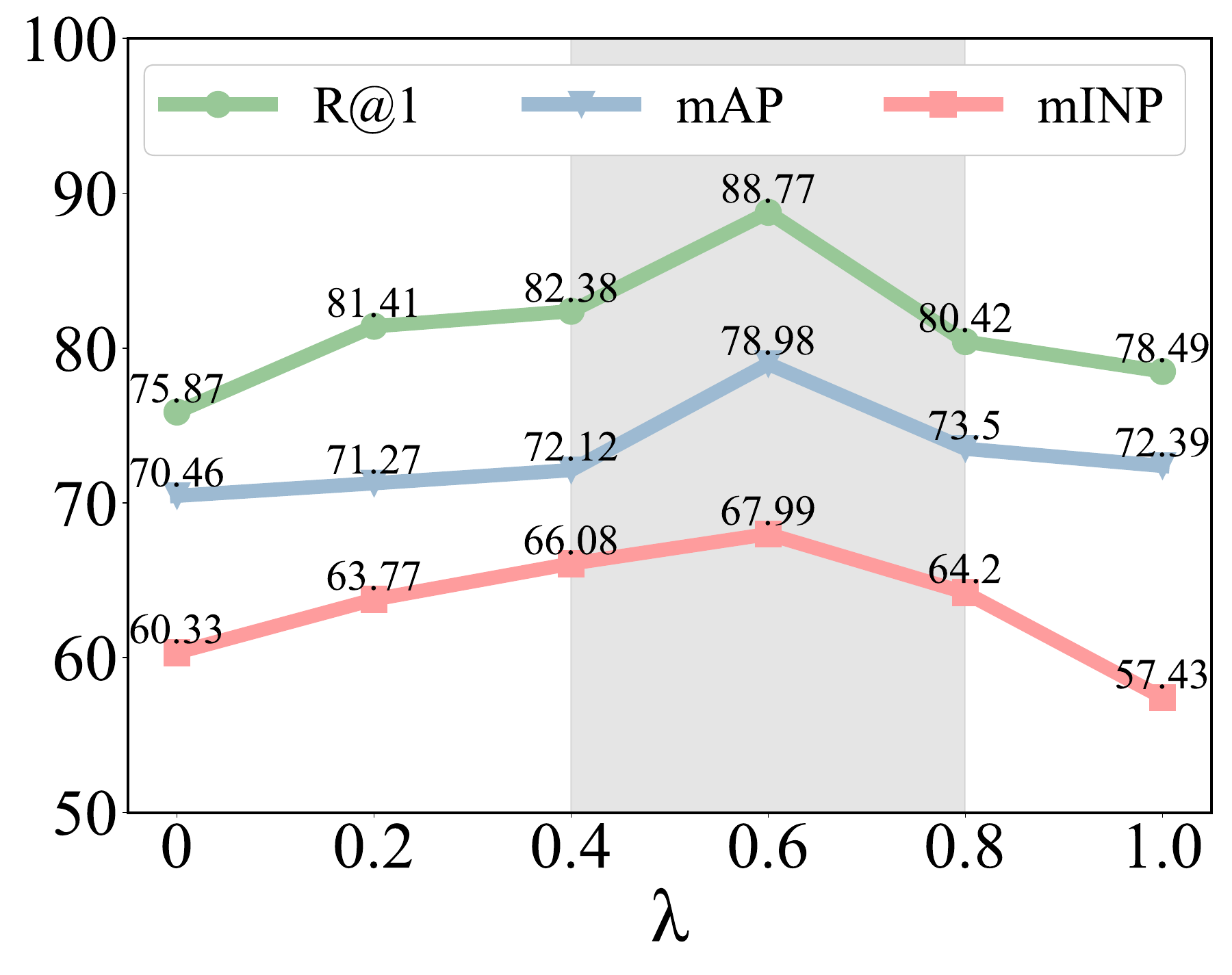}
        \caption*{(b) RegDB:V2T}
        \label{fig5b}
    \end{subfigure}
    \caption{\small The impact of parameter $\lambda$. The \textbf{\textcolor{gray}{gray shaded area}} represents the recommended parameter range for further fine-tuning, as suggested by the authors.}
    \label{fig5}
\end{figure}

As shown in~\Cref{fig5}, we analyze the performance variation of the model for different values of $\lambda$ within the range of $[0, 1]$. The sensitivity of $\lambda$ can vary depending on the characteristics of different datasets, but our results demonstrate that RoDE generally achieves the best performance when $\lambda$ is around $0.6$. Notably, when $\lambda = 0$ (i.e., without $\mathcal{L}_{ra}^{\alpha}$) or $\lambda = 1$ (i.e., without $\mathcal{L}_{ra}^{\beta}$), the model's performance significantly degrades. This highlights the critical role of balancing both discriminative learning and modality-invariant feature representation. The poor performance at the extremes of $\lambda$ underscores the necessity of jointly optimizing these two objectives to effectively address the challenges posed by noisy and incomplete data. 
\begin{table*}[t]
    \small 
    \centering
    \setlength{\tabcolsep}{1.5pt}
    \caption{\small Ablation studies of RoDE on SYSU-MM01, RegDB and LLCM datasets.}
    \setlength{\tabcolsep}{2.5mm}{
    \begin{tabular*}{0.96\textwidth}{@{\extracolsep{\fill}}c|l|ccc|ccc|ccc@{\extracolsep{\fill}}}
         \toprule[1.25 pt]
         \multirow{2}{*}{Order} & \multirow{2}{*}{Components} & \multicolumn{3}{c|}{\makecell[c]{SYSU-MM01: All Search}} & \multicolumn{3}{c|}{\makecell[c]{RegDB:V2T}} & \multicolumn{3}{c}{\makecell[c]{LLCM:V2T}} \\
         \cline{3-11}
         & & Rank-1 & mAP & mINP & Rank-1 & mAP & mINP & Rank-1 & mAP & mINP \\
         \midrule
         1 & RoDE w/o RAL ($\mathcal{L}_{ra}^{\alpha}$) & 54.17 & 52.96 & 32.11 & 75.87 & 72.46 & 60.33 & 33.72 & 34.18 & 21.72 \\
         2 & RoDE w/o RAL ($\mathcal{L}_{ra}^{\beta}$)  & 59.89 & 55.83 & 38.39 & 84.36 & 75.36 & 62.09 & 34.25 & 35.38 & 21.97 \\
         3 & RoDE w/o CCM (cross-model)           & 3.56  & 6.39  & 3.70 & 8.03  & 8.08  & 8.69 & 5.37  & 7.28  & 4.49  \\
         4 & RoDE w/o CCM (cross-modal)           & 44.31 & 43.90 & 29.83 & 52.88 & 45.98 & 29.25 & 30.18 & 31.94 & 18.44 \\
         5 & RoDE & \textbf{62.88}                & \textbf{57.91} & \textbf{43.04} & \textbf{88.77} & \textbf{78.98} & \textbf{67.99} & \textbf{35.13} & \textbf{37.44} & \textbf{23.04} \\
        \bottomrule[1.25 pt]
    \end{tabular*}}
    \label{tab3}
\end{table*}

\subsection{Ablation Studies}
\label{subsec:experiment6}
To address the three challenging issues arising from noise in pseudo-labels (i.e, noise overfitting, error accumulation, and noisy cluster correspondence), we designed three components, RAL, RDL, and CCM, respectively. The ablation studies are conducted to validate the effectiveness of them.

\subsubsection{Effectiveness of RAL} In this experiment, two variants are presented to study the effectiveness of our RAL: RoDE without $\mathcal{L}_{ra}^{\alpha}$ and RoDE without $\mathcal{L}_{ra}^{\beta}$, which is shown in \Cref{tab3}. The experimental results demonstrate that each component (i.e., $\mathcal{L}_{ra}^{\alpha}$ or $\mathcal{L}_{ra}^{\beta}$) contributes to the person re-identification performance. More specifically, RAL reduces over-optimization on low-reliability samples by using the robust loss function $\mathcal{L}_{ra}$, directly addressing noise interference. Furthermore, RAL adaptively ensures selective emphasis, effectively minimizing noise while preserving valuable information in mislabeled samples. Notably, compared to the absence of $\mathcal{L}_{ra}^{\beta}$, the lack of $\mathcal{L}_{ra}^{\alpha}$ often results in worse results. This phenomenon occurs because CCM establishes preliminary associations between cross-modal clusters with the same identity, enabling learning potential modality-invariant representations.

\subsubsection{Effectiveness of CCM} 
As shown in~\Cref{tab3}, removing CCM leads to suboptimal model performance, especially in the absence of cross-model alignment. The main reason may be that the lack of consistent correspondence leads to discontinuous and inconsistent supervision. Without cross-model consistency, the model suffers from interference caused by semantically inconsistent pseudo-labels generated by other models, leading to poor performance (i.e., Order \#3 in~\Cref{tab3}). In contrast, inter-modalities can maintain a certain degree of relevance during training, even without cross-modal alignment. This internal relevance helps integrate information and mitigate the impact of mismatches, thus avoiding significant performance degradation (i.e., Order \#4 in~\Cref{tab3}). Therefore, cross-model mismatch is a crucial issue affecting model performance and even making the model invalid.

\subsubsection{Effectiveness of RDL} 
We evaluate the effectiveness of RDL in capturing diverse information, as shown in ~\Cref{tab4}. Our method, which jointly trains and tests both models ($A+B$), outperforms independently trained models ($A$ and $B$) and collaborative models where only one model is used for testing. Specifically, $A+B(A)$ refers to collaborative training with model $A$ used for testing, and $A+B(B)$ refers to collaborative training with model $B$ used for testing. In contrast, our approach allows both models to benefit from each other’s predictions during training and testing, reducing error accumulation and leveraging complementary information.
\begin{table}[ht]
    \small 
    \centering
    \setlength{\tabcolsep}{4pt}
    \caption{\small Ablation studies on RDL.}
    \setlength{\tabcolsep}{1.8mm}{
    \begin{tabular*}{0.95\columnwidth}{@{\extracolsep{\fill}}l|ccc|ccc@{\extracolsep{\fill}}}
        \toprule[1.25 pt]
                & \multicolumn{3}{c|}{\makecell[c]{SYSU-MM01: All Search}} & \multicolumn{3}{c}{\makecell[c]{RegDB: V2T}} \\
                \cline{2-7}
                 & Rank-1 & mAP & mINP & Rank-1 & mAP & mINP \\
         \midrule
         A       & 56.81 & 52.77 & 35.92 & 70.41 & 65.77 & 57.92 \\
         B       & 56.66 & 53.02 & 36.88 & 71.12 & 66.02 & 58.88 \\
         A+B (A) & 60.23 & 54.86 & 40.18 & 84.32 & 73.86 & 63.18 \\
         A+B (B) & 59.97 & 54.27 & 37.44 & 83.97 & 73.27 & 63.44 \\
         A+B     & \textbf{62.88} & \textbf{57.91} & \textbf{43.04} & \textbf{88.77} & \textbf{78.98} & \textbf{67.99} \\
         \bottomrule[1.25 pt]
    \end{tabular*}}
    \label{tab4}
\end{table}

\begin{figure}[ht]
    \centering    
    \begin{subfigure}[c]{0.49\linewidth}
        \centering
        \includegraphics[width=\columnwidth]{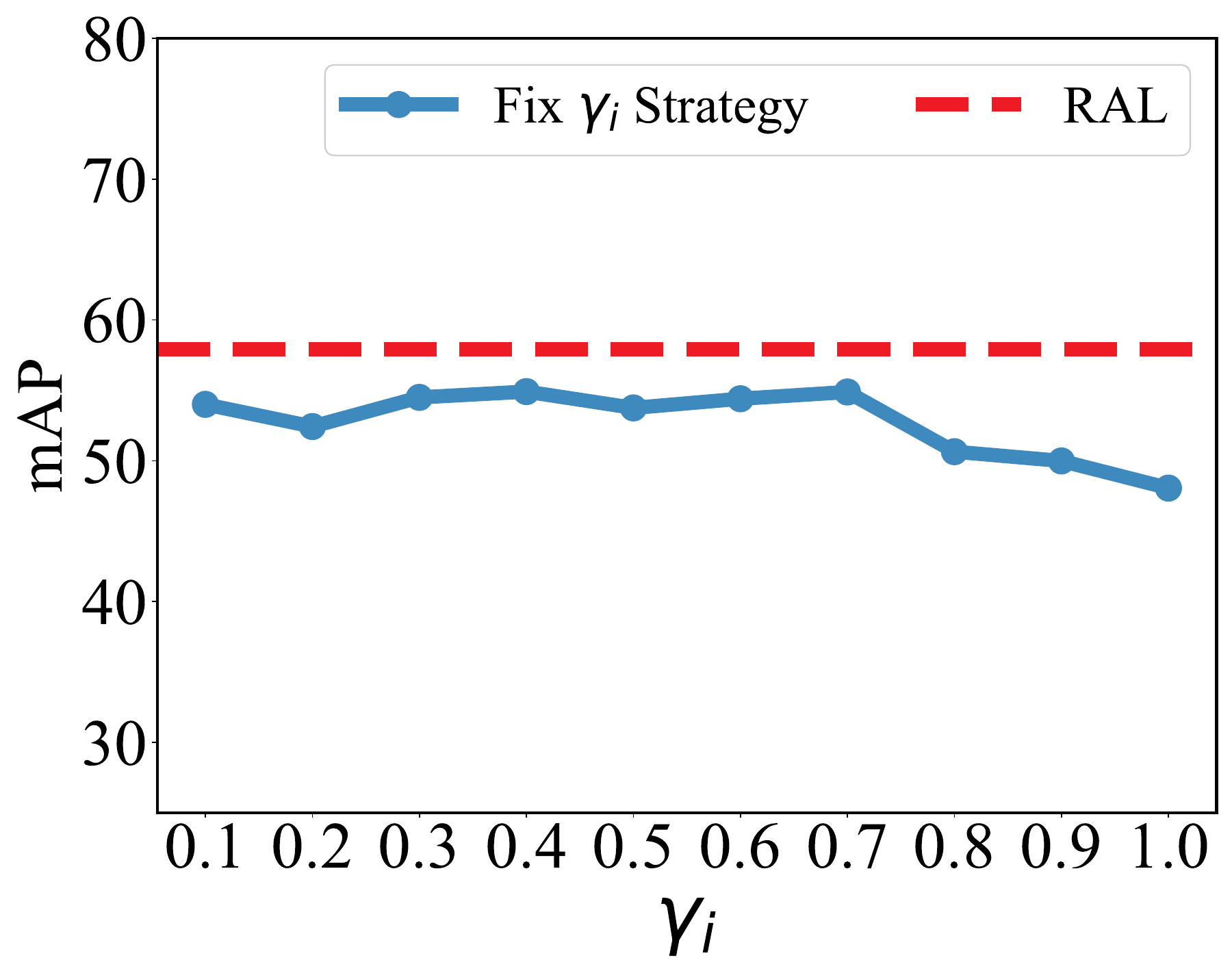}
        \caption*{(a) SYSU-MM01:All Search}
        \label{fig6a}
    \end{subfigure}
    \begin{subfigure}[c]{0.49\linewidth}
        \centering
        \includegraphics[width=\columnwidth]{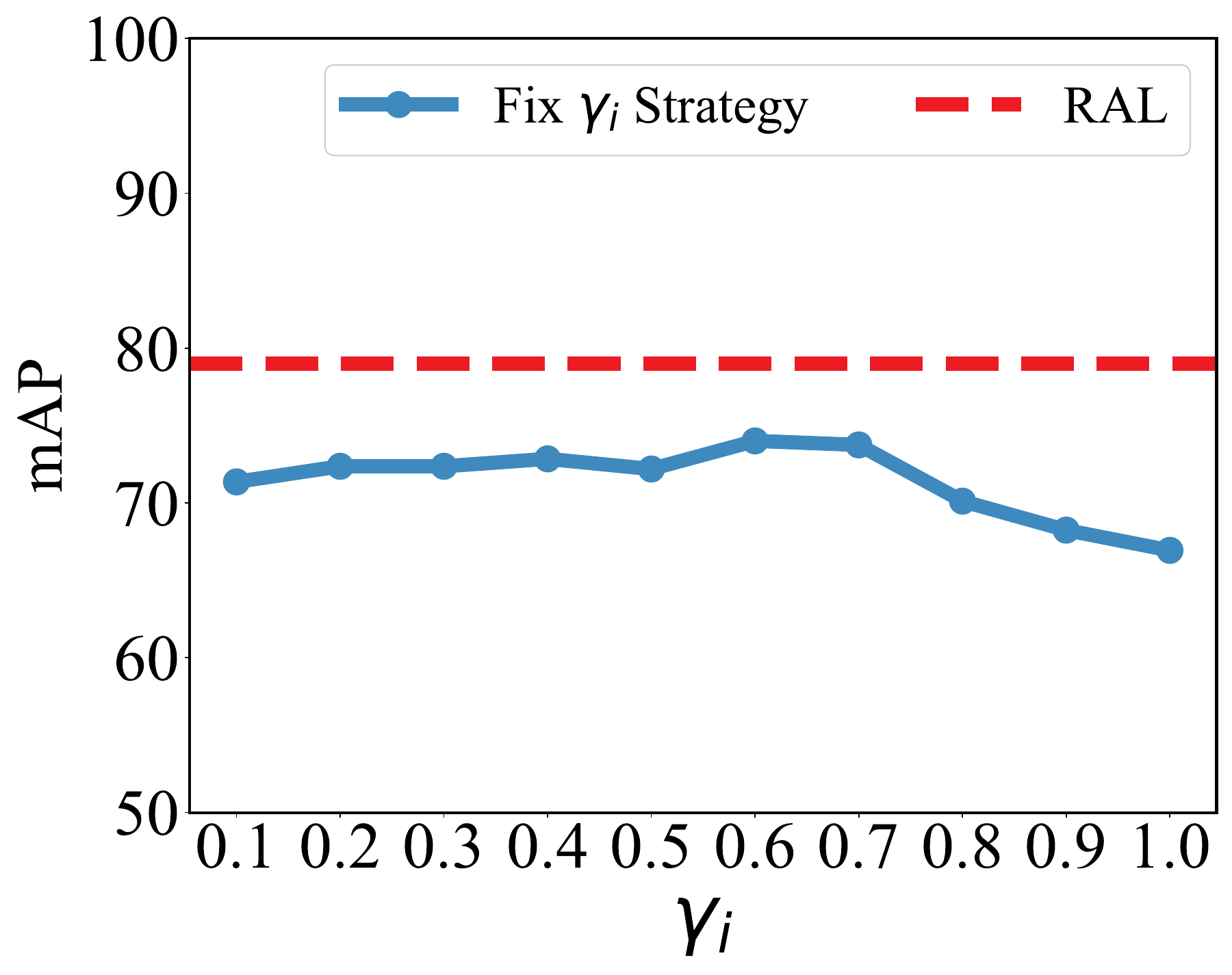}
        \caption*{(b) RegDB:V2T}
        \label{fig6b}
    \end{subfigure}
    \caption{\small The impact of the parameter $\gamma_i$ and the advantages of the adaptive strategy in RAL. The \textcolor[rgb]{0.243,0.541,0.745}{blue} points indicate results with a fixed value of $\gamma_i$, while the \textcolor[rgb]{0.929,0.110,0.141}{red} line represents the results of the RAL, which serves as the upper bound for the fixed $\gamma_i$ strategy.}
    \label{fig6}
\end{figure}

\subsubsection{The Beneficial Effects of Self-adaption Selecting~\texorpdfstring{$\gamma_i$}{}} 
We conducted a series of experiments to demonstrate the advantages of the adaptive optimization strategy in RAL. Specifically, we compared RAL with strategies using fixed parameter $\gamma_i$, where $\gamma_i$ ranges in $(0, 1]$, which is shown in \Cref{fig6}. From the figure, one can see that the result of RAL serves as an upper bound (the red line) compared to the results achieved with various fixed parameters $\gamma_i$, demonstrating that the fixed parameter $\gamma_i$ limits the ability of adaptive optimization for the clean and noisy samples. In other words, RAL effectively mitigates performance degradation by adaptively reducing the emphasis on mislabeled samples.

\subsection{Visualization Analysis}
\label{subsec:experiment7}
We conduct a detailed visualization comparing RoDE with the most competitive baselines GUR~\cite{yang2023towards} and DOTLA~\cite{cheng2023unsupervised}.

\subsubsection{t-SNE Visualization} 
\begin{figure*}[ht]
    \centering        
    \begin{subfigure}[c]{0.325\linewidth}
        \centering
        \includegraphics[width=\columnwidth]{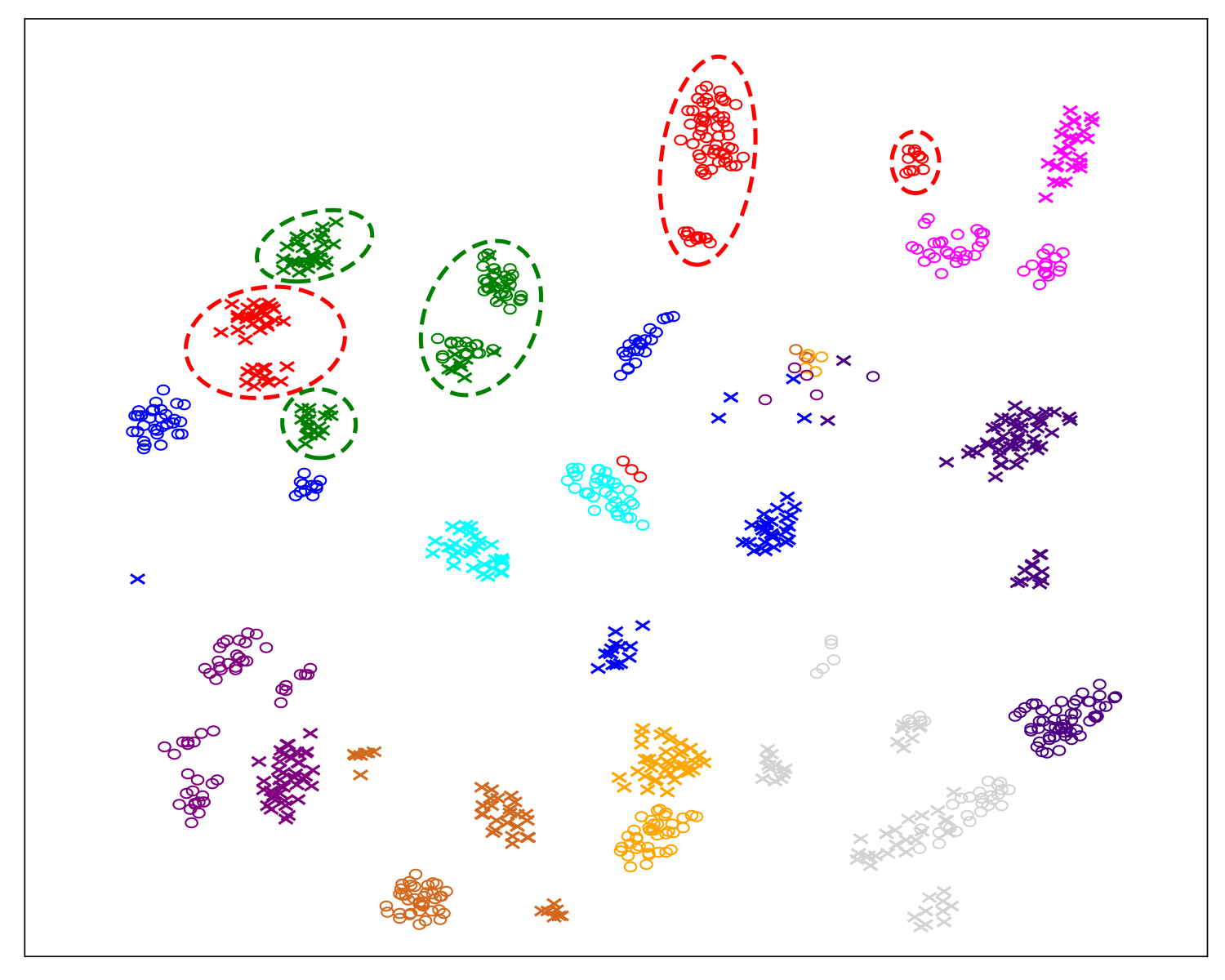}
        \caption*{(a) GUR~\cite{yang2023towards}}
        \label{fig7a}
    \end{subfigure}
    \begin{subfigure}[c]{0.325\linewidth}
        \centering
        \includegraphics[width=\columnwidth]{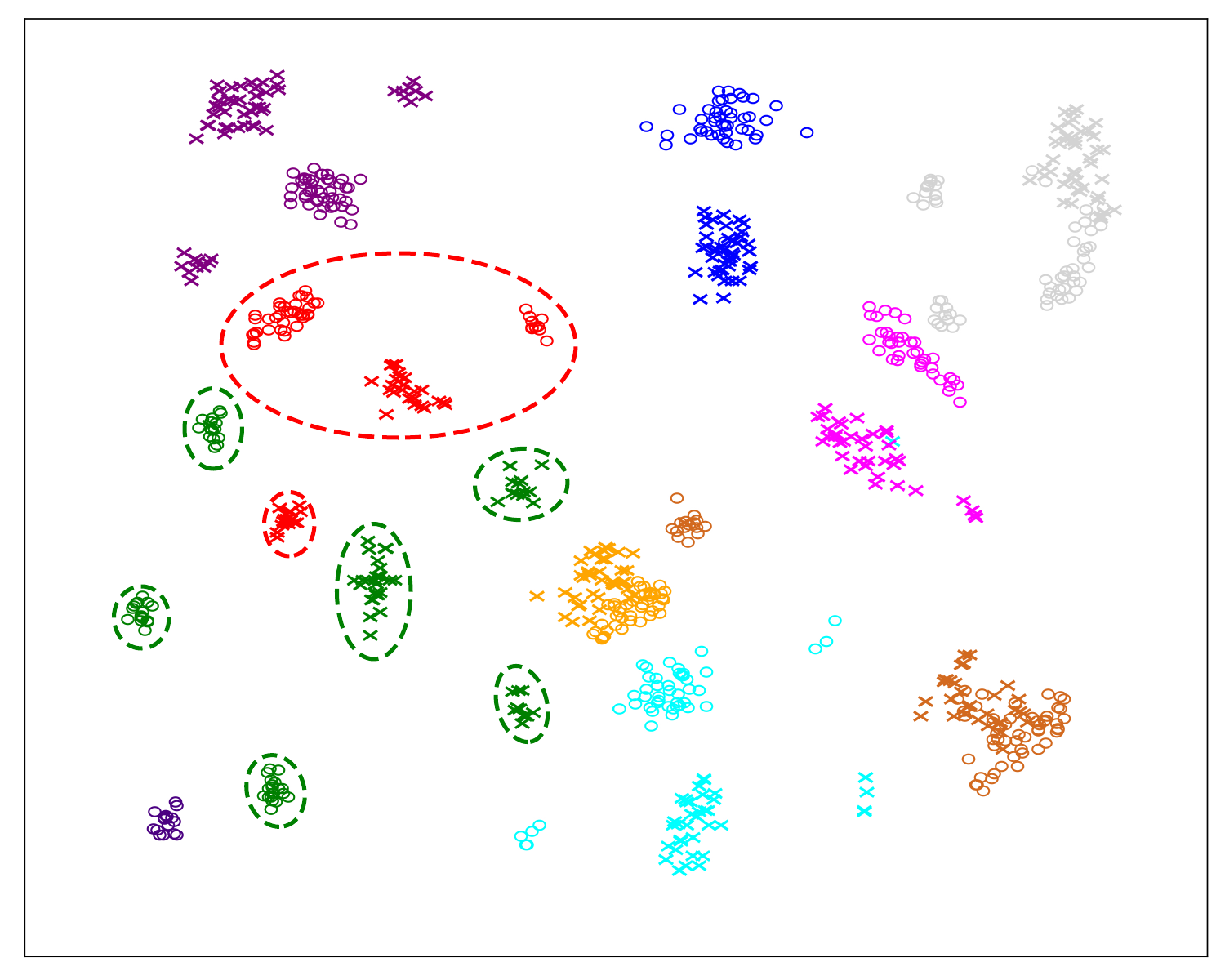}
        \caption*{(b) DOTLA~\cite{cheng2023unsupervised}}
        \label{fig7b}
    \end{subfigure}
    \begin{subfigure}[c]{0.325\linewidth}
        \centering
        \includegraphics[width=\columnwidth]{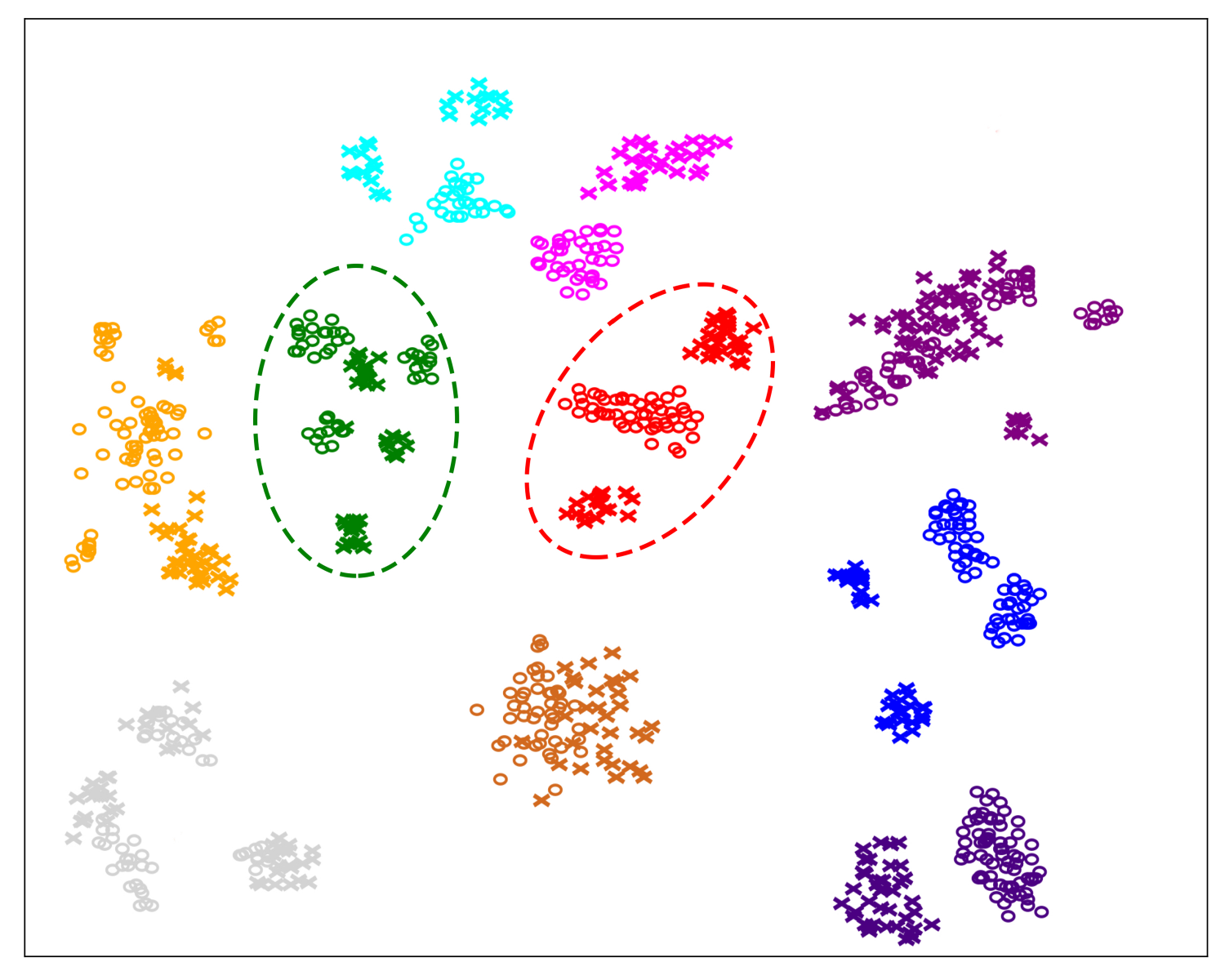}
        \caption*{(c) RoDE}
        \label{fig7c}
    \end{subfigure}
    \caption{\small The t-SNE plot for 10 randomly selected identities from SYSU-MM01 is presented, with $\circ$ representing visible modality and $\times$ representing infrared modality.}
    \label{fig7}
\end{figure*}

We plot the t-SNE map feature distribution of 10 randomly selected identities from SYSU-MM01 in~\Cref{fig7}. We observe that GUR~\cite{yang2023towards} and DOTLA~\cite{cheng2023unsupervised} fail to come together pedestrian images with the same identities. For example, in~\Cref{fig7} (a) and (b), the samples marked in red and green do not flock together within the same dashed ellipse. This issue likely arises from their inability to robustly handle noise interference in pseudo-labels, which results in the distribution of some samples in the feature space being shifted by label noise. In contrast, RoDE (i.e.,~\Cref{fig7} (c)) demonstrates robustness against pseudo-label noise, producing a consistent understanding of images with the same identity in the feature space even under noisy label conditions.

\subsubsection{Visualization on the Qualitative Results}
\begin{figure}[ht]
    \centering    
    \begin{subfigure}[c]{0.0586\linewidth}
        \centering
        \includegraphics[width=\columnwidth]{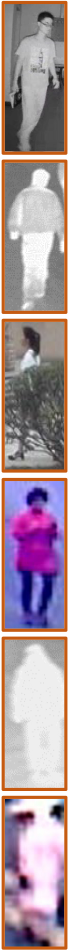}
        \caption*{ }
        \label{fig8q}
    \end{subfigure}
    \begin{subfigure}[c]{0.29\linewidth}
        \centering
        \includegraphics[width=\columnwidth]{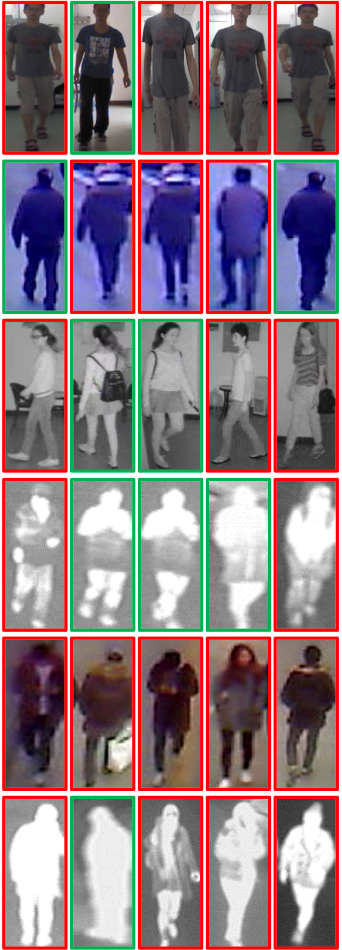}
        \caption*{(a) GUR~\cite{yang2023towards}}
        \label{fig8a}
    \end{subfigure}
    \begin{subfigure}[c]{0.29\linewidth}
        \centering
        \includegraphics[width=\columnwidth]{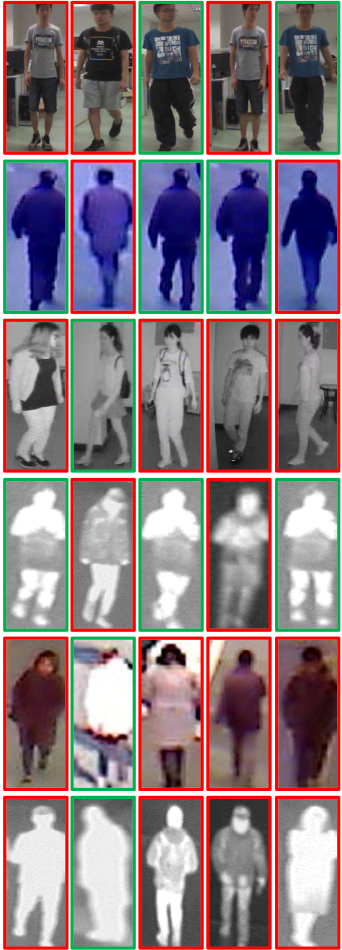}
        \caption*{(b) DOTLA~\cite{cheng2023unsupervised}}
        \label{fig8b}
    \end{subfigure}
    \begin{subfigure}[c]{0.29\linewidth}
        \centering
        \includegraphics[width=\columnwidth]{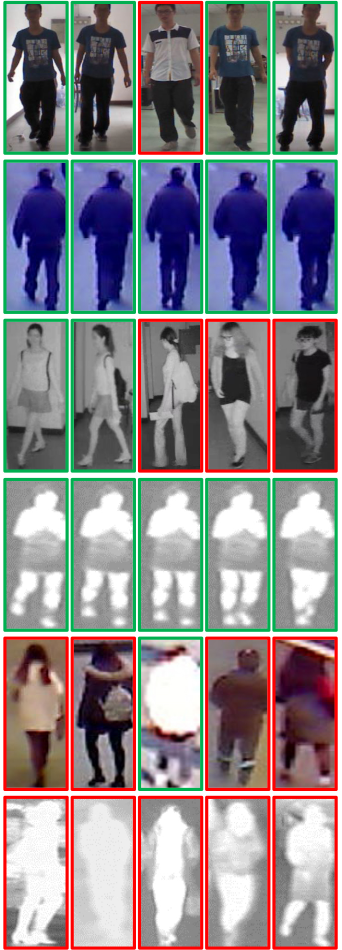}
        \caption*{(c) RoDE}
        \label{fig8c}
    \end{subfigure}
    
    \caption{\small Some person re-identification results of (a) GUR~\cite{yang2023towards}, (b) DOTLA~\cite{cheng2023unsupervised}, and (c) RoDE. Each row presents a query image of a person (marked with an \textbf{\textcolor{orange}{orange}} bounding box) on the left, with the retrieved images highlighted in \textbf{\textcolor{green}{green}} bounding boxes denoting correct matches and those in \textbf{\textcolor{red}{red}} indicating incorrect matches. The results are arranged in descending order. The first four cases show successful outcomes, while the last two represent failures.}
    \label{fig8}
\end{figure}

We further evaluated the qualitative results of RoDE with the benchmark methods. The qualitative results are presented by retrieving the Top-5 gallery images with the highest similarity scores for each query image, as shown in~\Cref{fig8}. RoDE provides more stable matching results compared to other competitive methods. Notably, RoDE selects accurate results even for the challenging query image with severe occlusion (i.e., the first-row case), demonstrating its ability to handle complex scenarios. However, RoDE faces inevitable failure when the query image is severely blurred and unclear (i.e., the third-row and fourth-row case). This is because, in cases of significant image degradation, the blurred visual information makes it difficult for the model to extract effective features, thereby affecting the matching accuracy.

\subsection{Robustness Analysis}
\label{subsec:experiment8}
To verify the robustness of the proposed RoDE against pseudo-label noise (PLN), we conduct detailed experiments and analyses focusing on three challenges: noisy overfitting, error accumulation, and noisy cluster correspondence. While these three types of noise are interrelated and can all significantly impact model performance, no single type can be considered more critical than the others.

\subsubsection{Robustness Analysis on Noisy Overfitting} 
\begin{figure}[t]
    \centering        
    \begin{subfigure}[c]{0.49\linewidth}
        \centering
        \includegraphics[width=\columnwidth]{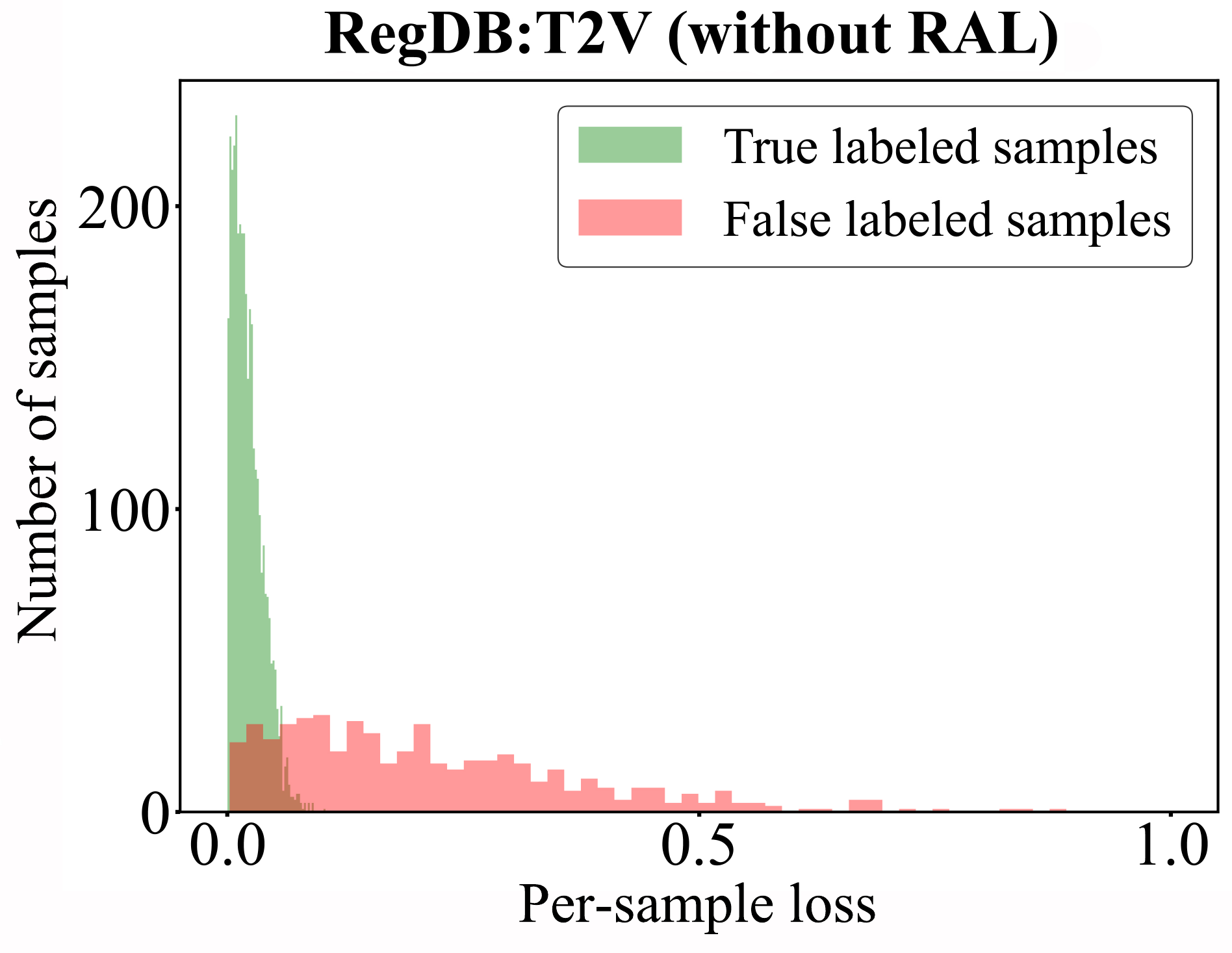}
        \caption*{(a) without RAL}
        \label{fig9a}
    \end{subfigure}
    \begin{subfigure}[c]{0.49\linewidth}
        \centering
        \includegraphics[width=\columnwidth]{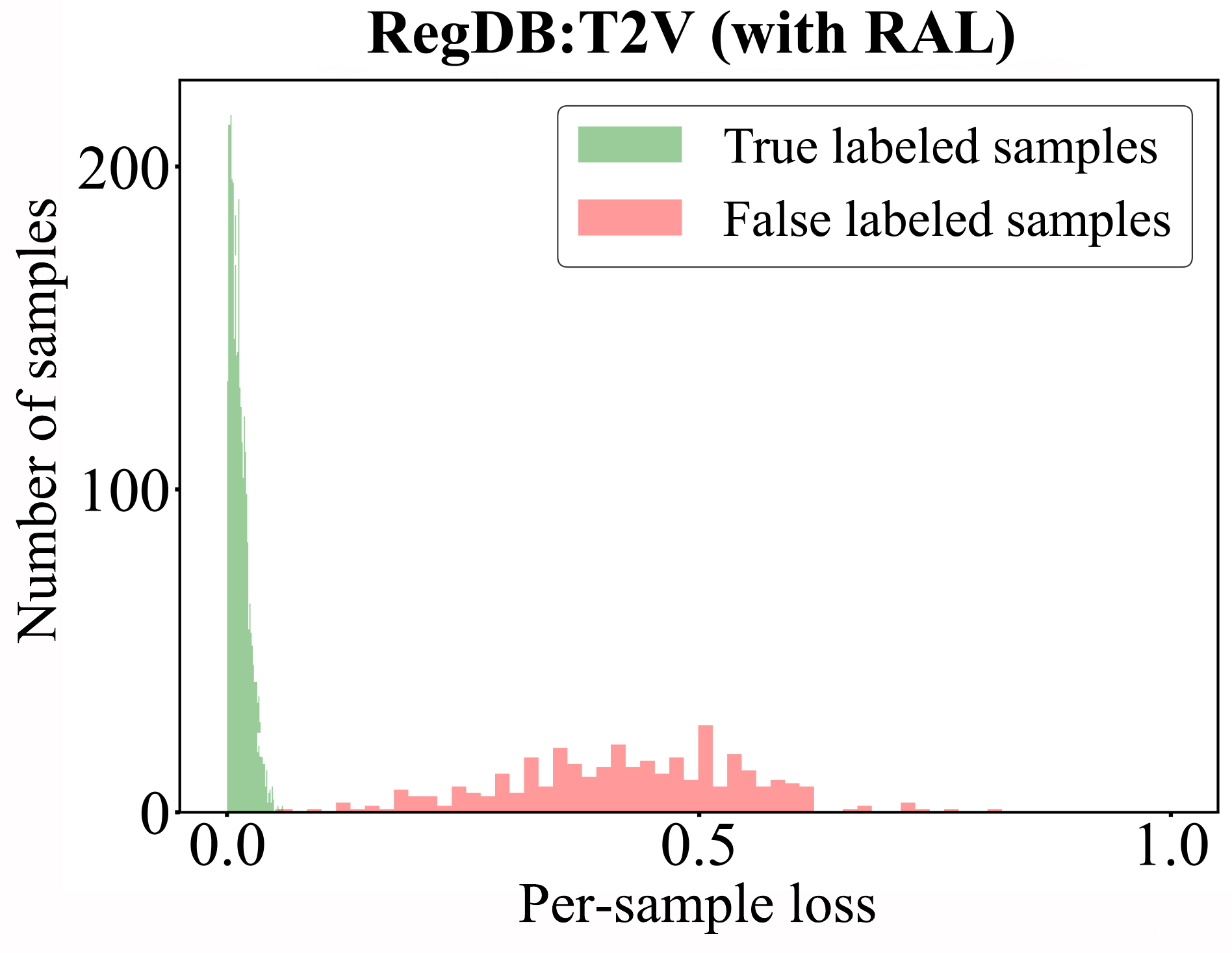}
        \caption*{(b) with RAL}
        \label{fig9b}
    \end{subfigure}
    
    \begin{subfigure}[c]{0.49\linewidth}
        \centering
        \includegraphics[width=\columnwidth]{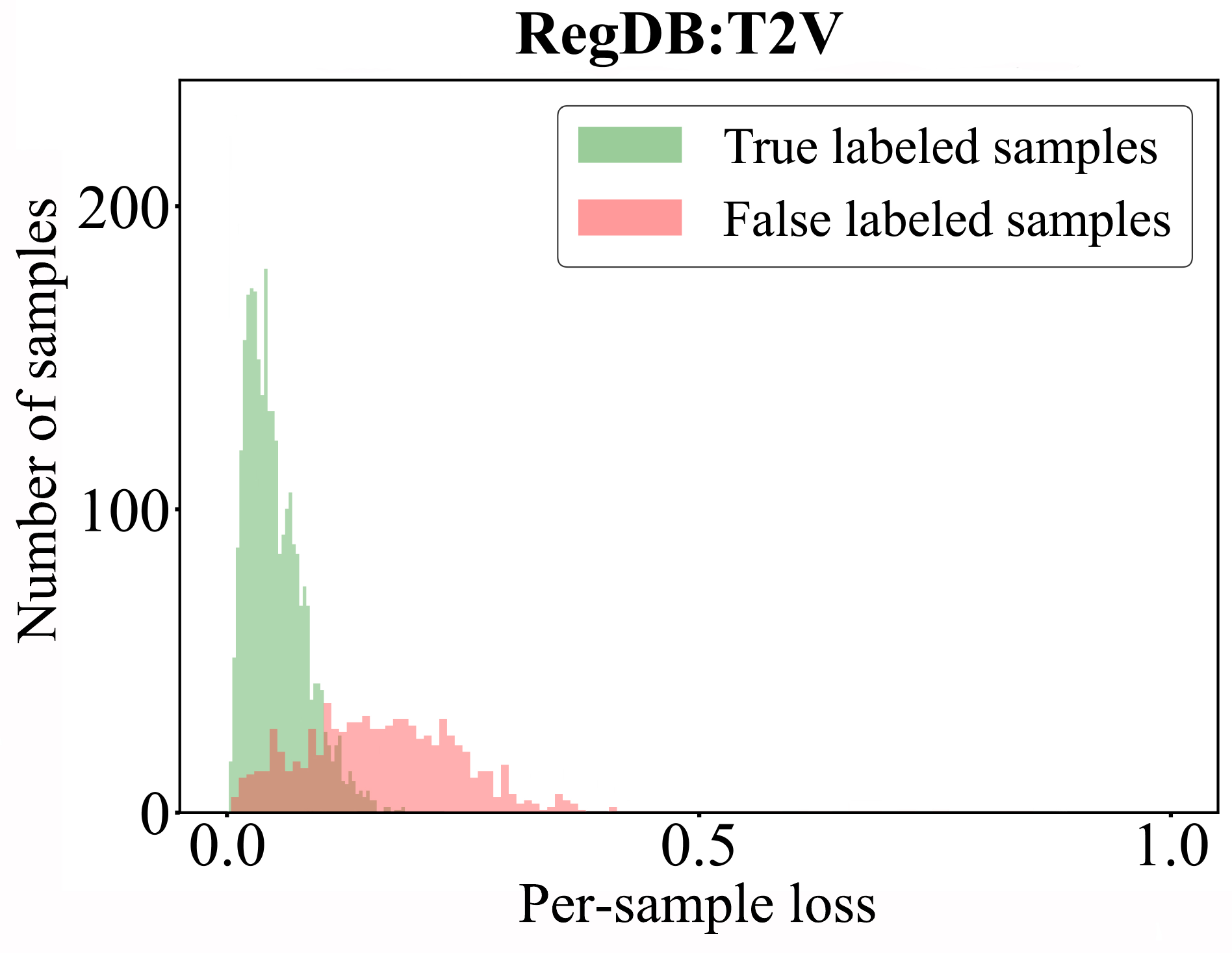}
        \caption*{(c) Single Model}
        \label{fig9c}
    \end{subfigure}
    \begin{subfigure}[c]{0.49\linewidth}
        \centering
        \includegraphics[width=\columnwidth]{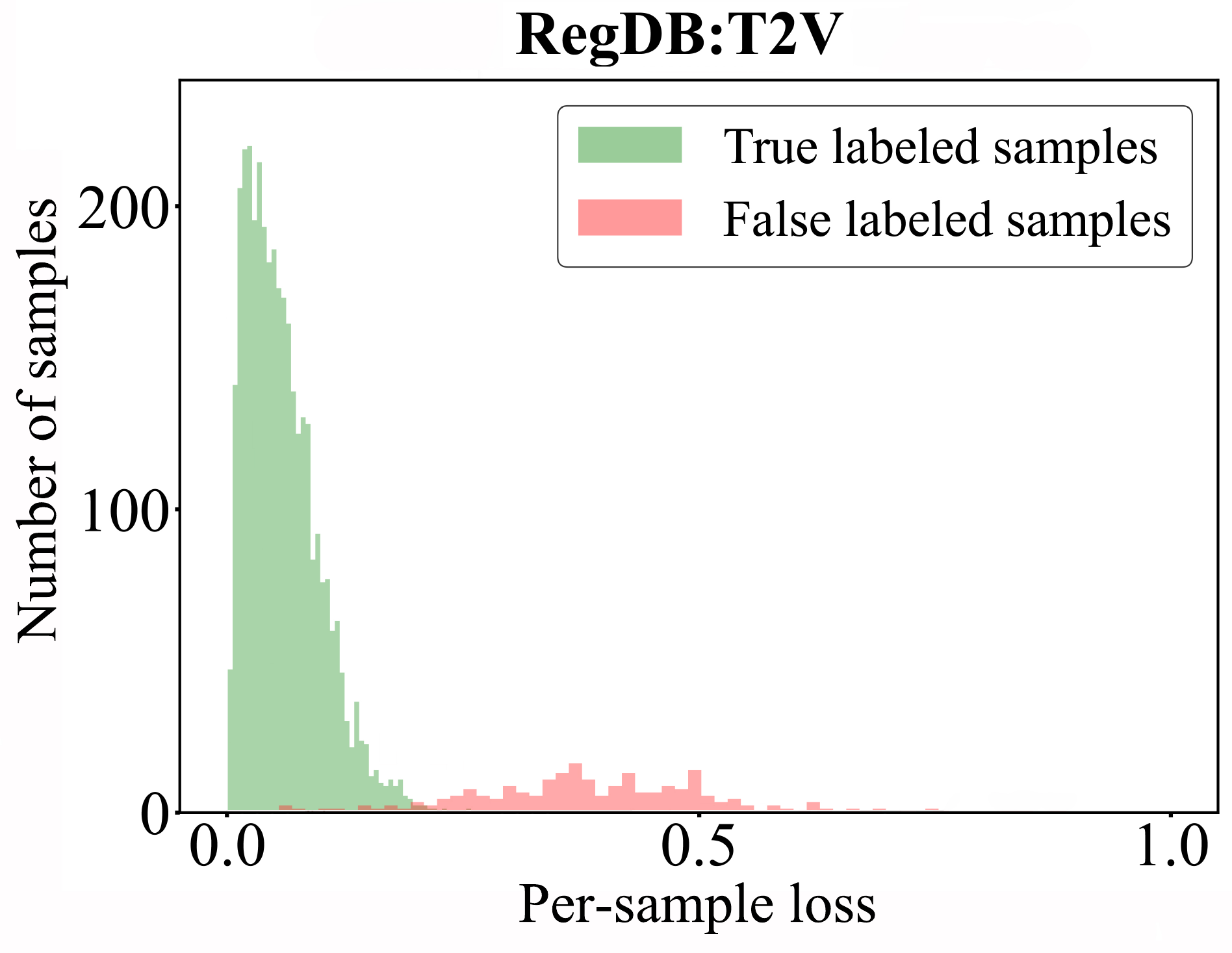}
        \caption*{(d) Dual Models}
        \label{fig9d}
    \end{subfigure}
    
    \begin{subfigure}[c]{0.49\linewidth}
        \centering
        \includegraphics[width=\columnwidth]{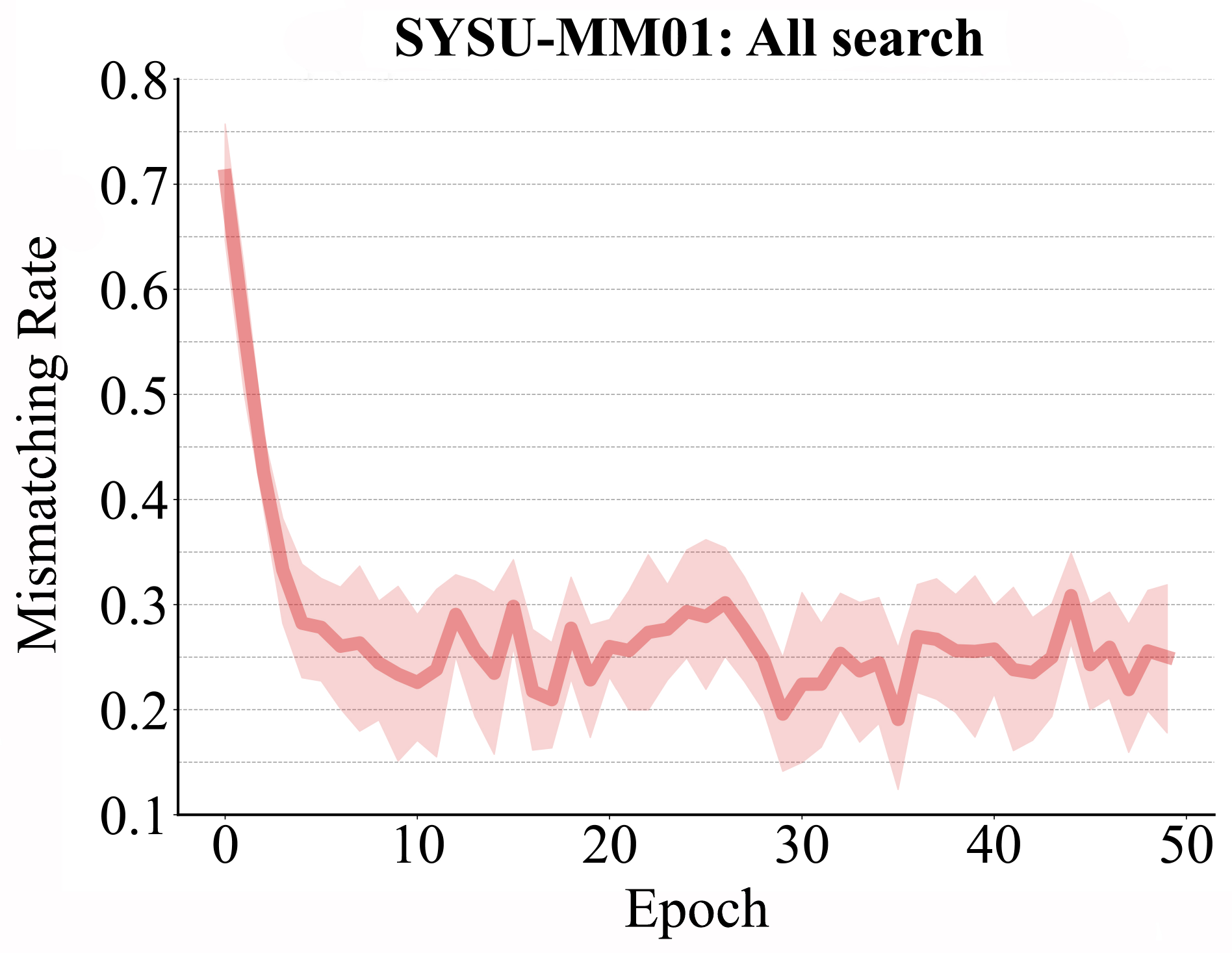}
        \caption*{(e) Cross-modal Mismatching}
        \label{fig9e}
    \end{subfigure}
    \begin{subfigure}[c]{0.49\linewidth}
        \centering
        \includegraphics[width=\columnwidth]{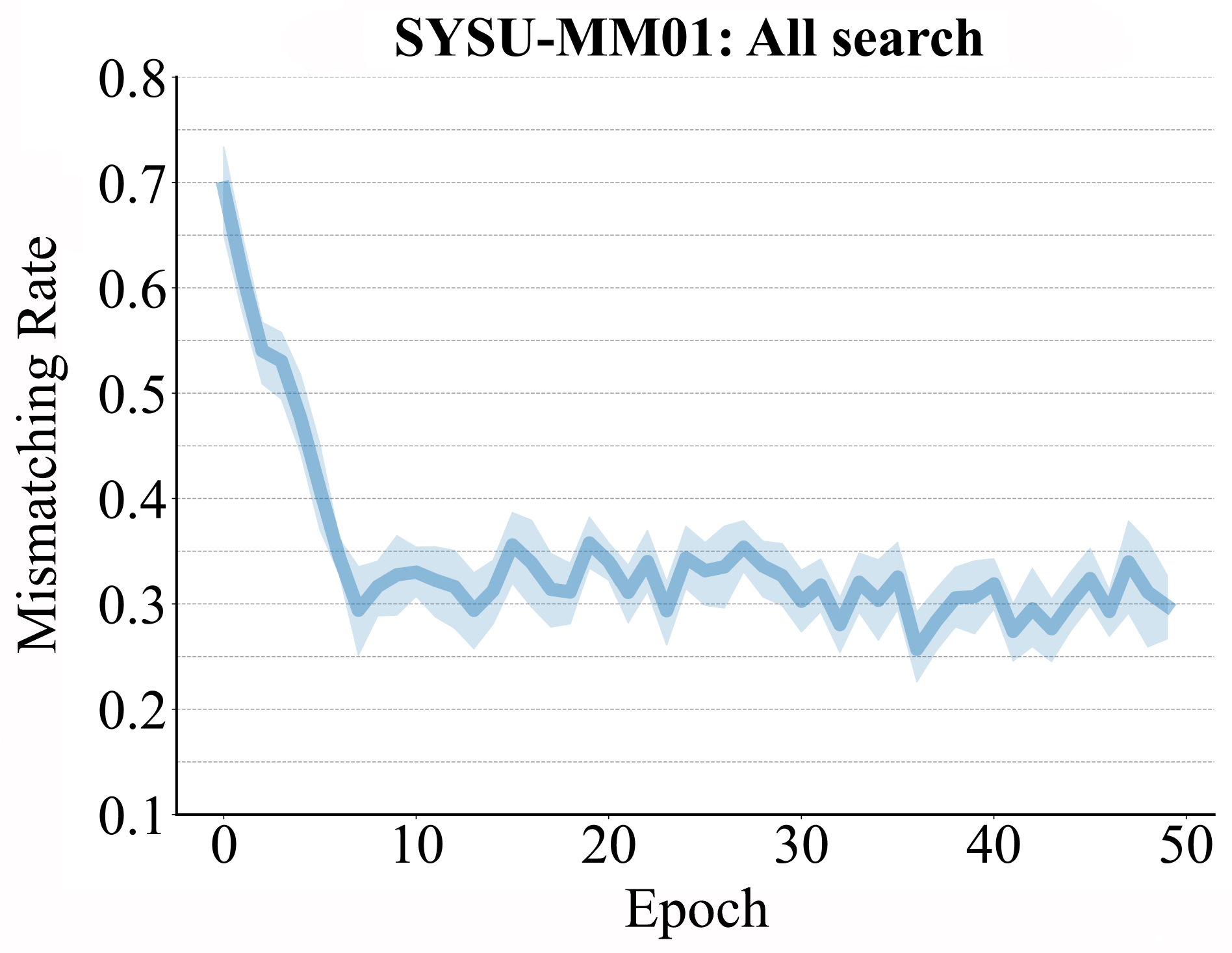}
        \caption*{(f) Cross-model Mismatching}
        \label{fig9f}
    \end{subfigure}
    \caption{\small Robustness analysis of RoDE.}
    \label{fig9}
\end{figure}

To illustrate the issue of noisy overfitting, we analyze the sample loss distribution, as illustrated in~\Cref{fig9} (a) and (b). When RAL is absent, many clean samples and noisy labeled samples appear simultaneously near the low loss value, indicating severe overfitting to the noisy samples. In contrast, with the introduction of RAL, the loss distribution exhibits a clear separation between clean and noisy samples. This improvement is because RAL can reduce the attention to noisy samples through adaptive optimization, thus preventing the training process from being dominated by pseudo-label noise.

\subsubsection{Robustness Analysis on Error Accumulation}
The mutual learning mechanism in the RDL strategy reduces error accumulation by alternating training between two models and sharing pseudo-labels. It also prevents both models from becoming overly biased towards incorrect pseudo-labels, mitigating error accumulation. This alternating process ensures that both models benefit from each other’s predictions, improving generalization and reducing susceptibility to errors from relying on a single model. Moreover, the RDL framework uses a dynamic weighting mechanism RAL to adjust the influence of pseudo-labels based on their reliability, ensuring that errors do not dominate the learning process. To display error accumulation, we have counted the loss distributions for each sample in the infrared modality. These distributions are obtained through training a single model (i.e.,~\Cref{fig9} (c)) and dual models (i.e.,~\Cref{fig9} (d)), respectively. In the case of single models, noisy and clean samples intermingle due to significant error accumulation, as evidenced by the overlapping colored areas. This overlap indicates that the model struggles to differentiate between noisy and clean samples, leading to imprecise predictions. In contrast, dual models with RoDE effectively alleviate this issue, producing a clear separation between noisy and clean samples.

\subsubsection{Robustness Analysis on Noisy Cluster Correspondence} 
We observe that both cross-model and cross-modal cluster mismatches decrease progressively during training and eventually stabilize within a certain range, as illustrated in~\Cref{fig9} (e) and (f). The shaded area indicates the standard deviation. This demonstrates that CCM can mitigate biases introduced by noise or inconsistent correspondences, helping the model maintain a more accurate optimization path. Unfortunately, while CCM reduces most of the matching errors, a mismatching rate of 15\% to 35\% still persists, which underscores the need for further refinement in the matching process.

\section{Conclusion and Future Works}
\label{sec:conclusion}
This paper proposes a novel learning paradigm, RoDE, for UVI-ReID that simultaneously addresses three key challenges: noisy overfitting, error accumulation, and noisy cluster correspondence. To mitigate noisy overfitting, RoDE employs a pivotal RAL to dynamically and adaptively reduce the emphasis on noisy samples. It also alternates training between two individual models, thereby maintaining diversity and avoiding error accumulation. Additionally, RoDE incorporates the CCM to establish reliable alignment across distinct modalities and different models by leveraging cross-cluster similarities. Numerous experiments demonstrate the excellent performance of our proposed method. In the future, we plan to extend RoDE to tackle additional challenges in VI-ReID, particularly noise filtering techniques and domain adaptation, to better handle the variability of real-world scenarios. 

\bibliography{references}{}
\bibliographystyle{ieeetr}

\end{document}